\documentclass{article}

\usepackage[final]{corl_2021} % Uncomment for the camera-ready ``final'' version.

\usepackage{booktabs}
\usepackage{caption}
\usepackage{subcaption}
\usepackage{amsmath}
\usepackage{graphicx}
\usepackage{amssymb}
\usepackage{wrapfig}
\usepackage{bibunits}
\usepackage{chngpage}
\usepackage{etoc} 
\usepackage{enumitem}
\usepackage{dcolumn}
\usepackage{bookmark}
\newcolumntype{d}[1]{D{,}{\,\pm\,}{#1}}

\renewcommand{\contentsname}{Supplement Contents}
\etocsettocstyle{\section*{\contentsname}}{}%

\newcommand{\beginsupplement}{%
    \setcounter{table}{0}
    \renewcommand{\thetable}{S\arabic{table}}%
    \setcounter{figure}{0}
    \renewcommand{\thefigure}{S\arabic{figure}}%
    \setcounter{section}{0}
 }

\usepackage{color, soul} % This allows us to highlight text that you want your co-authors to review
\renewcommand\hl[1]{#1} % Uncomment this to turn off highlighting 

\DeclareRobustCommand{\hlc}[1]{{\sethlcolor{cyan}\hl{#1}}}
\renewcommand\hlc[1]{#1}

\DeclareMathOperator*{\argmax}{arg\,max}

\newcommand{\fig}[1]{Fig.~\ref{#1}}
\newcommand{\appfig}[1]{Appendix Fig.~\ref{#1}}
\newcommand{\tbl}[1]{Table~\ref{#1}} 
 
\newcommand{\sect}[1]{Sec.~\ref{#1}} 
\newcommand{\appsect}[1]{Appendix Sec.~\ref{#1}}
 
\newcommand{\etal}[0]{{\em et al.~}} 
\newcommand{\eg}[0]{{\em e.g.,~}} 
\newcommand{\ie}[0]{{\em i.e.,~}}

\title{FabricFlowNet: Bimanual Cloth Manipulation with a Flow-based Policy}

\author{
  Thomas Weng, Sujay Bajracharya, Yufei Wang, Khush Agrawal, and David Held\\
  Robotics Institute, Carnegie Mellon University, USA\\
  \texttt{\{tweng, sbajrach, yufeiw2, khusha, dheld\}@andrew.cmu.edu}
}

\begin{document}
\maketitle

%===============================================================================

\begin{bibunit}
\vspace{-2em}
\begin{abstract}
We address the problem of goal-directed cloth manipulation, a challenging task due to the deformability of cloth. 
Our insight is that optical flow, a technique normally used for motion estimation in video, can also provide an effective representation for corresponding cloth poses across observation and goal images.
We introduce FabricFlowNet (FFN), a cloth manipulation policy that leverages flow as both an input and as an action representation to improve performance.
FabricFlowNet also elegantly switches between dual-arm and single-arm actions based on the desired goal.
We show that FabricFlowNet significantly outperforms state-of-the-art model-free and model-based cloth manipulation policies. 
We also present real-world experiments on a bimanual system, demonstrating effective sim-to-real transfer.  
Finally, we show that our method generalizes when trained on a single square cloth to other cloth shapes, such as T-shirts and rectangular cloths.
Video and other supplementary materials are available at: \mbox{\href{https://sites.google.com/view/fabricflownet}{https://sites.google.com/view/fabricflownet}}.
\end{abstract}

\keywords{deformable object manipulation, optical flow, bimanual manipulation} 

\section{Introduction}
\label{sec:intro}

Cloth manipulation has a wide range of applications in domestic and industrial settings. However, it has posed a challenge for robot manipulation: compared to rigid objects, fabrics have a higher-dimensional configuration space, can be partially observable due to self-occlusions in crumpled configurations, and do not transform rigidly when manipulated. 
Early approaches for cloth manipulation relied on scripted actions; these policies are typically slow and do not generalize to arbitrary cloth goal configurations~\cite{5509439, 7589002, 6095109}.

Recently, learning-based approaches have been explored for cloth manipulation~\cite{Hoque-RSS-20, seita_fabrics_2020, yan2020learning, nair2017combining, seita2020learning}, including model-free reinforcement learning to obtain a  policy~\cite{lee2020learning, Wu-RSS-20}. For a cloth manipulation policy to be general to many different objectives, it must receive a representation of the current folding objective.
% In order to obtain a general cloth-folding policy, the policy must receive a representation of the folding objective. 
A standard approach for representing a goal-conditioned policy is to input an image of the current cloth configuration together with an image of the goal~\cite{lee2020learning, seita2020learning}. 

We will show a number of downsides to such an approach when applied to cloth manipulation.  First, the policy must learn to reason about the relationship between the current observation and the goal, while also reasoning about the action needed to obtain that goal.  These are both difficult learning problems; requiring the network to reason about them jointly exacerbates the difficulty. 
Additionally, previous work has used reinforcement learning (RL) to try to learn such a policy~\cite{lee2020learning, Wu-RSS-20}; however, a reward function is a fairly weak supervisory signal, which makes it difficult to learn a complex cloth manipulation policy.  
Finally, while many desirable folding actions are more easily and accurately manipulated with bimanual actions, previous learning-based methods for goal-conditioned cloth manipulation have been restricted to single-arm policies.

\begin{figure}[t]
    \centering
    \includegraphics[width=0.99\textwidth]{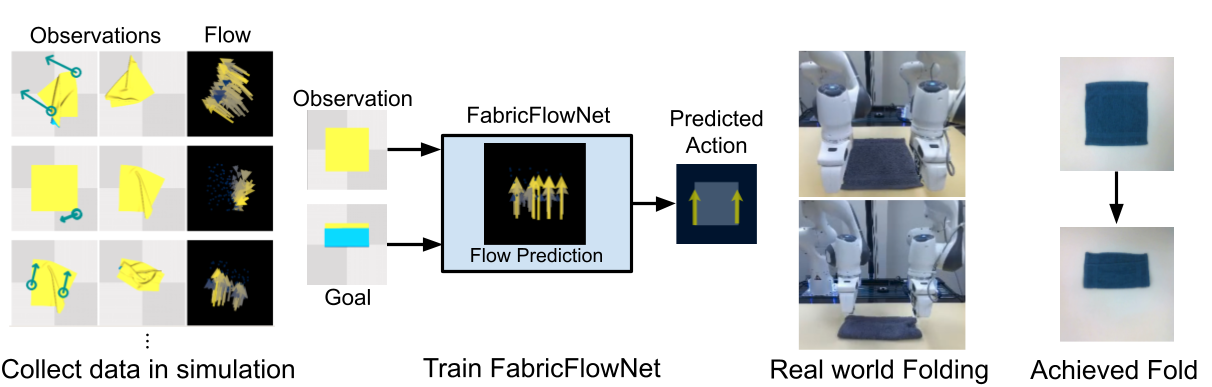}
    \caption{FabricFlowNet (FFN) overview. We collect a dataset of random actions and ground truth flow to train FFN. FFN learns to predict flow and uses it as both an input and action representation in a manipulation policy. FFN successfully performs single and dual-arm folding in the real world.}
    \label{fig:overview}
    % https://docs.google.com/drawings/d/1sVuqntF3_sAndjLo1EDNTXaVxzSf2ko8ca7nOvRDCzw/edit?usp=sharing
    \vspace{-1em}
\end{figure}

In this paper, we introduce FabricFlowNet (FFN), a goal-conditioned policy for bimanual cloth manipulation that uses optical flow to improve policy performance (see \fig{fig:overview}).
Optical flow has typically been used for video-related tasks such as object tracking and estimating camera motion. 
We demonstrate that flow can also be used in the context of policy learning for cloth manipulation; we use an optical flow-type network to estimate the relationship between the current observation and a sub-goal. We use flow in two ways: first, as an input representation to our policy; second, after estimating the pick points for a pick-and-place policy, we query the flow image to determine the place actions. Our method is learned entirely with supervised learning\hl{, leveraging ground truth particles from simulation}. Our method learns purely from random actions without any expert demonstrations during training and without reinforcement learning. 

Our learned policy can perform bimanual manipulation and switches easily between dual and single-arm actions, depending on what is most suitable for the desired goal. 
Our approach significantly outperforms our best efforts to extend recent single-arm cloth manipulation approaches to bimanual manipulation tasks~\cite{Hoque-RSS-20, lee2020learning}. 
We present experiments on a dual-arm robot system and in simulation evaluating our method's cloth manipulation performance.
FabricFlowNet outperforms state-of-the-art model-based and model-free baselines, and 
we provide extensive ablation experiments to demonstrate the importance of each component of our method to the achieved performance. Our method also generalizes with no additional training to other cloth shapes and colors.
This paper contributes:
\begin{itemize}[leftmargin=1em]
    \item A novel flow-based approach for learning goal-conditioned cloth manipulation policies that can perform dual-arm and single-arm actions.
    \item \hl{A test suite for benchmarking goal-conditioned cloth folding algorithms encompassing and expanding on goals used in previous literature\mbox{~\cite{Hoque-RSS-20, lee2020learning, ganapathi2020learning}}; we perform extensive experiments using this test suite to evaluate FabricFlowNet (FFN), baselines\mbox{~\cite{Hoque-RSS-20, lee2020learning}}, and ablations, demonstrating that FFN outperforms previous approaches.}
    \item \hl{Experiments to demonstrate that FFN generalizes to other cloth colors and shapes, even without training on such variations.}
\end{itemize}

\section{Related Work}
\label{sec:related}

\textbf{Bimanual Manipulation}.
A large body of research exists on dual-arm, or bimanual, manipulation~\cite{SMITH20121340}.
Dual-arm systems allow for more complex behaviors than single-arm systems
at the cost of greater planning complexity~\cite{edsinger2007two, shauri2011assembly}, leading to research on closed kinematic chain planning~\cite{suarez2015using, bordalba2020randomized}, composable skill learning~\cite{NEURIPS2020_18a010d2, chitnis2020efficient}, and rewarding synergistic behavior~\cite{chitnis2020intrinsic}.
Prior work has also explored bimanual cloth manipulation~\hl{\mbox{\cite{Sanchez2018RoboticMA}}, including establishing a diverse set of benchmark tasks~\mbox{\cite{8957044}}}.
Cloth manipulation is a highly underactuated task, and bimanual manipulation enables controlling multiple cloth points~\cite{borras2020grasping}. 
A common approach for cloth flattening is to lift a cloth with one arm and regrasp it with the other arm until it reaches the flattened configuration~\cite{Li_ICRA2015, 5980327, 5509439, 7589002, 6095109}.
Previous work in this direction uses hard-coded policies~\cite{5509439, 7589002, 6095109}, whereas we learn to achieve arbitrary folded configurations.
Tanaka~\etal\cite{8276243} learn bimanual actions for goal-conditioned folding, using a voxel-based dynamics model to predict how actions will change the cloth state.
\hl{However, optimizing this dynamics model can slow down inference time compared to our model-free approach.}
Dynamic bimanual manipulation has also been explored in simulation from ground-truth keypoints~\cite{jangir2020dynamic} and for unfolding cloth in the real world~\cite{ha2021flingbot}; we perform real-world bimanual folding using depth image observations.

\textbf{Learning for Cloth Manipulation.}
Prior works have proposed various hand-defined representations for cloth manipulation, such as parameterized shape models~\cite{Miller2012AGA} or binary occupancy features~\cite{6942687}. 
Recent approaches use contrastive learning to learn pixel-wise latent embeddings for cloth~\cite{ganapathi2020learning, chi2021garmentnets}. \hl{Both contrastive learning~\mbox{\cite{ganapathi2020learning}} and goal-conditioned transporter networks~\mbox{\cite{seita2020learning}} have been applied to imitate expert demonstrations. Our approach doesn't require expert actions, just sub-goal states provided at test-time to define the task. In contrast to these previous representations, our method uses a flow-based representation, which we found to significantly outperform previous methods for goal-based cloth manipulation.}

\hl{Other approaches have applied policy learning techniques to single-arm cloth smoothing~\mbox{\cite{Wu-RSS-20,seita_fabrics_2020}}. 
In contrast, we learn a policy that performs either single and dual-arm cloth manipulation; further, our focus is on goal-conditioned cloth folding, rather than smoothing.}
For cloth manipulation, Lee~\etal\cite{lee2020learning} learns a model-free value function, \hl{but is limited by its discrete action space, and further, they do not use a flow-based representation, which we show leads to large benefits.}
Prior methods for learning goal-conditioned policies have used self-supervised learning to learn an inverse dynamics model for rope~\cite{nair2017combining, pathakICLR18zeroshot} \hlc{but such approaches have not been demonstrated for cloth manipulation.} 
\hl{Lippi~\mbox{\etal\cite{lippi2020latent}} plan cloth folding actions in latent space, but do not demonstrate generalization to unseen cloth shapes.
Other papers use an online simulator~\mbox{\cite{Li_ICRA2015}}, or learn a cloth dynamics model in latent space~\mbox{\cite{yan2020learning}}, pixel-space~\mbox{\cite{Hoque-RSS-20}}, or over a graph of keypoints~\mbox{\cite{ma2021learning}}.
Unlike these model-based methods, our method is model-free and does not require online simulation or time-expensive CEM planning, leading to much faster inference.}
Further, we compare our approach to state-of-the-art approaches for cloth manipulation~\cite{lee2020learning,Hoque-RSS-20} and show significantly improved performance.

\textbf{Optical Flow for Policy Learning.}
Optical flow is the task of estimating per-pixel correspondences between two images, typically for video-related tasks such as object tracking and motion estimation. 
State-of-the-art approaches use convolutional neural networks (CNN) to estimate flow~\cite{dosovitskiy2015flownet, ilg2017flownet, teed2020raft}.
Optical flow between successive observations has previously been used as an input representation to capture object motion for peg insertion~\cite{dong2021tactile} or dynamic tasks~\cite{amiranashvili2018motion}.
Within the domain of cloth manipulation, Yamazaki~\etal\cite{doi:10.5772/61930} similarly use optical flow on successive observations to identify failed actions. 
We use flow not to represent motion between successive images, but to correspond the cloth pose between observation and goal images, and to determine the placing action for folding.
Argus~\etal\cite{argus2020flowcontrol} use flow in a visual servoing task to compute residual transformations between images from a demonstration trajectory and observed images.
In contrast, we learn a policy with flow to determine what cloth folding actions to take, not how to servo to a desired pose. 
\section{Learning a Goal-Conditioned Policy for Bimanual Cloth Manipulation}
\label{sec:approach}

\subsection{Problem Definition}
\label{sec:problemdef}

Our objective is to enable a robot to perform cloth folding manipulation tasks.
Let each task be defined by a sequence of sub-goal observations $\mathcal{G}: \{x_1^g, x_2^g, \dots, x_N^g\}$, each of which can be achieved by a single (possibly bimanual) pick-and-place action from the previous sub-goal. 
We require sub-goals, rather than a single goal, because a folded cloth can be highly self-occluded such that a single goal observation fails to describe the full goal state.
Defining a task using a sequence of sub-goals is found in other recent work~\cite{pathakICLR18zeroshot}.
Similar to prior work~\cite{pathakICLR18zeroshot, nair2017combining}, even if the sub-goals are obtained from an expert demonstration, we nonetheless do not assume access to the expert actions; this is a realistic assumption if the sub-goals are obtained from visual observations of a human demonstrator.

We assume that the agent does not have access to the sub-goal sequence $\mathcal{G}$ during training that it must execute during inference.
Thus, the agent must learn a general goal-conditioned policy $a_t = \pi(x_t, \mathcal{G})$, where $x_t$ is the current observation of the cloth and $a_t \in \mathcal{A}$ is the action selected by the policy.  
In our approach, we input each sub-goal $x_i^g$ sequentially to our policy: 
$a_t = \pi(x_t, x_i^g)$. A goal recognizer~\cite{pathakICLR18zeroshot} can also be used to decide which sub-goal observation to input at each timestep. For convenience, we will interchangeably refer to $x_i^g$ as a goal or sub-goal.

\begin{figure}[t!]
    \centering
    \begin{subfigure}[t]{0.37\textwidth}
        % https://docs.google.com/drawings/d/1cILZz3C5FKlw9CdDzYq7JwpPi3NV77DsM8X3z_vM9pU/edit
        % \includegraphics[width=\textwidth,natheight=726,natwidth=1126]{figures/naive_system.png}
        \includegraphics[width=\textwidth]{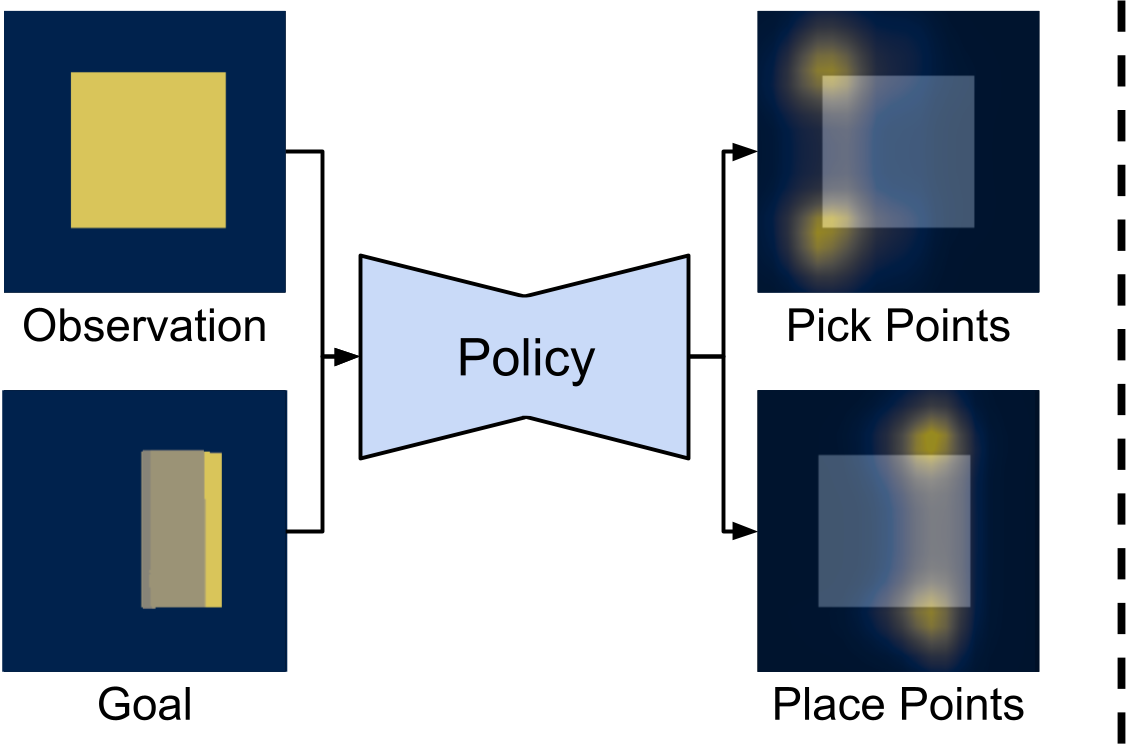}
        \caption{Naive system}
        \label{fig:naive_system}
    \end{subfigure}
    \begin{subfigure}[t]{0.62\textwidth}
        % https://docs.google.com/drawings/d/1fDd5eR2rAjfBIAJJUU-_LEeaTutvB4AolNanwHeUYfs/edit
        % \includegraphics[width=\textwidth,natheight=204,natwidth=512]{figures/system.png}
        \includegraphics[width=\textwidth]{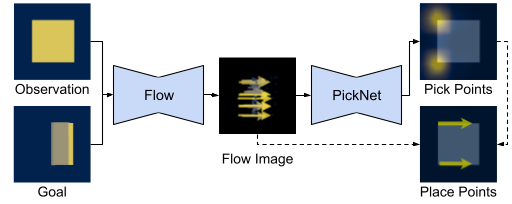}
        \caption{FabricFlowNet (FFN)}
        \label{fig:system}
    \end{subfigure}
    \caption{(\subref{fig:naive_system}) A naive approach to goal-conditioned policy learning is to input observation and goal images directly to the policy and predict the action. (\subref{fig:system}) FabricFlowNet separates representation learning from policy learning; it first estimates the correspondence between the observation and goal as a flow image. The flow is then used as the input to PickNet for pick point prediction, and as a way to compute place points without requiring additional learning.}
    \vspace{-2.5em}
\end{figure}

\subsection{Overview}

A common approach for a goal-conditioned policy is to input the current observation $x_t$ and the goal observation $x_i^g$ directly 
into to a neural network representation of a policy~\cite{pathakICLR18zeroshot, lee2020learning} or a Q-function~\cite{Wu-RSS-20, lee2020learning}. However, the network must reason simultaneously about the relationship between the observation and the goal, as well as the correct action to achieve that goal. Our first insight is that we can improve performance by separating these two components: we will learn to reason about the relationship between the observation and the goal, and separately use this relationship to reason over actions. Specifically, we represent this relationship using a ``flow image'' $f$, which indicates the correspondence between the current observation $x_t$ and sub-goal $x_i^g$. Thus we propose using the flow image $f$ as an improved input representation of the policy, rather than directly inputting the observation $x_t$ and goal observation $x_i^g$. 

Our second insight is that we can also use flow in the output representation of the policy. 
We use a pick and place action space;
prior methods that learn pick and place policies for deformable object manipulation 
predict place points using the policy network, 
either explicitly~\cite{seita_fabrics_2020, seita2020learning, yan2020learning,nair2017combining, pathakICLR18zeroshot,Wu-RSS-20} or implicitly by transforming the inputs to a Q-function~\cite{lee2020learning}.
Instead, we simplify the problem by leveraging flow: our policy network only learns to predict the \emph{pick} points. For the place point, we query the flow image $f$ for the flow vector starting at the predicted pick location, and use the endpoint of that vector as the place point. 

We demonstrate that using flow in the two ways described above for our policy achieves significantly improved performance compared to prior work. Furthermore, our approach extends naturally to dual-arm manipulation, allowing us to easily transition between single and dual-arm actions.

A schematic overview of our system can be found in \fig{fig:system}.
We first compute the flow $f$ between the current observation $x_t$ and goal $x_i^g$.
Next, we input the flow $f$ to a policy network (PickNet), which outputs pick points $p_i$.
We then query the flow image $f(p_i)$ to determine the place points for each robot arm. 
Further details of our approach are described below.

\subsection{Estimating Flow between Observation and Goal Images}
\label{sec:flow}

We learn flow to use it as an input representation to our pick prediction network, and as an action representation for computing place points.
Given an observed depth image $x_t$ and desired goal depth image $x_i^g$, we estimate the flow $f=(f^1, f^2)$, mapping each pixel $(u, v)$ in $x_t$ to its corresponding coordinates $(u', v') = (u + f^1(u), v + f^2(v))$ in $x_i^g$.  
This task formulation differs from standard optical flow tasks as the input image pairs $(x_t, x_i^g)$ are not consecutive images from video frames.

To capture the complex correspondences between $x_t$ and $x_i^g$, we train a convolutional neural network to estimate the flow image $f$ (see Appendix for details). 
The training loss we use to supervise the network is endpoint error (EPE), the standard error for optical flow estimation. 
EPE is the Euclidean distance between the predicted flow vectors $f$ and the ground truth $f^*$, averaged over all pixels: $\mathcal{L}_\text{EPE} = \frac{1}{N} \sum^N_{i=1} \| f^* - f \|_2$.
We use a cloth simulation to collect training examples with ground truth flow.
The simulator provides the ground-truth correspondence between the particles of the cloth in different poses.
The simulation cloth particles are not as dense as the depth image pixels; as a result, we only have ground-truth flow supervision for a sparse subset of the pixels that align with the cloth particles. Thus, we mask the loss to only supervise the flow for the pixels that align with the location of the cloth particles. We train the flow network using data collected from random actions.
See \sect{sec:implementation} for more details on the simulator, data collection, and network training. 

\begin{figure}[t!]
    \centering
    % https://docs.google.com/drawings/d/1A1N8SH6aWVA4W-8jLNag2UVL6_nndRLUqJk3tCkpaLk/edit
    % \includegraphics[width=0.99\textwidth,natwidth=2404,natwidth=712]{figures/picknet.png}
    \includegraphics[width=0.99\textwidth]{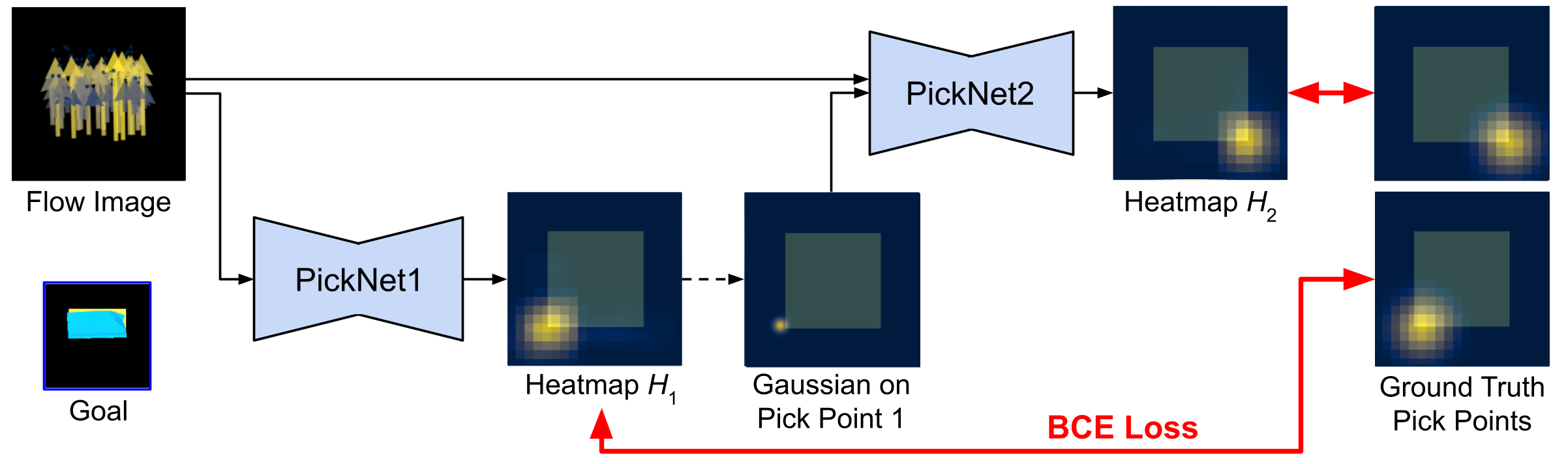}
    \caption{PickNet architecture. We utilize a two-network architecture for bimanual manipulation, where the second pick point is conditioned on the prediction of the first pick point.}
    \label{fig:picknet}
    \vspace{-1em}
\end{figure}

\subsection{Learning to Predict Pick Points}
\label{sec:picknet}
Our bimanual action space $\mathcal{A}$ consists of actions $a = (p_1, p_2, q_1, q_2)$, where $p$ and $q$ are the pick and place points respectively, paired according to the subscripts. 
We train a neural network called PickNet to estimate the pick points $p_1, p_2$.
Crucially, the input to PickNet is a flow image $f$, estimated between the current depth image $x_t$ and the desired goal depth image $x_i^g$, as described in the previous section.
The flow image indicates, for each pixel $(u,v)$ in the current observation, the location $f(u,v)$ that the pixel has moved to in the goal observation. 
Our flow network (\sect{sec:flow} above) reasons about the observation-goal relationship, so that the policy network (PickNet) only needs to reason about the action, specifically the two pick points ($p_1$, $p_2$); computing the place points is described in \sect{sec:place}.

For dual-arm actions, the pick points must be estimated conditionally, as the location of pick point $p_1$ on the cloth influences the optimal location of pick point $p_2$, and vice versa. 
To decouple this conditional estimation problem, we propose a two-network architecture, PickNet1 and PickNet2, to estimate the pick points (see \fig{fig:picknet}).
This architecture was inspired by Wu~\etal\cite{Wu-RSS-20}, which used two networks for pick-conditioned placing; we instead use two networks to condition dual-arm picking.
PickNet1 is a fully convolutional network that receives flow image $f$ as input and outputs a single heatmap $H_1$ estimating the optimal pick points for arm 1. We compute the first pick point as $p_1 = \argmax_{p} H_1(p)$.
The second network, PickNet2, predicts the second arm's pick point $p_2$ conditioned on $p_1$; PickNet2 takes as input both the flow image $f$ and an additional image with a 2D Gaussian centered on $p_1$, and is otherwise identical to PickNet1.
PickNet2 outputs heatmap $H_2$, from which we compute the second pick point: $p_2 = \argmax_{p} H_2(p)$. 
The two-network architecture decouples the conditionally dependent pick point predictions and does not require us to resort to heuristics to extract two pick points from a single heatmap. 
We refer to PickNet1 and PickNet2 together as ``PickNet."

To train PickNet, we collect a dataset of random actions (see \sect{sec:implementation} for details) and record the current observation $x_t$, the bimanual action $a = (p_1, p_2, q_1, q_2)$, and the next observation $x_{t+1}$. We also estimate the flow $f$ from $x_t$ to $x_{t+1}$, \hl{as explained} in \sect{sec:flow}.
We create ground truth pick heatmaps $H_i^*$ for arm $i$ using the recorded random action $a$, by placing a 2D Gaussian $\mathcal{N}(p_i, \sigma)$ on each ground truth pick location $p_i$.
We then supervise PickNet using the binary cross-entropy (BCE) loss between predicted heatmaps $H_1, H_2$ and ground truth heatmaps $H_1^*, H_2^*$. However, it might be unclear to the network which pick point should be output by PickNet1 and which should be output by PickNet2.  
We compute the loss for both possible correspondences and use the minimum:
\begin{equation}
   \begin{aligned}
    l_\text{BCE}(H_i, H_j, H_i^*, H_j^*) = \text{BCE}(H_i, H_i^*) + \text{BCE}(H_j, H_j^*)\\
  \mathcal{L}_\text{Pick} = \min[ l_\text{BCE}(H_1, H_2, H_1^*, H_2^*), l_\text{BCE}(H_2, H_1, H_1^*, H_2^*)]
  \label{eq:lpick}
  \end{aligned}
\end{equation}
At inference time, PickNet outputs the pick points $p_1, p_2$, computed from the argmax of $H_1, H_2$ respectively, as described above.

\subsection{Estimating the Place Points from Flow}
\label{sec:place}

After estimating the pick points $p_1, p_2$ from flow, the remaining step to predict a bimanual pick and place action $a = (p_1, p_2, q_1, q_2)$ is to estimate the place points $q_1, q_2$.
A straightforward approach would be to train the network to predict place points $q_1, q_2$, similar to the pick points $p_1, p_2$ as described above (see \fig{fig:naive_system}).
Instead, our approach uses the flow image to find the place points, so that the place points do not have to be learned separately. 

Our approach makes the assumption that, to achieve a desired subgoal configuration, the point picked on the cloth should be moved to its corresponding position in the goal image (which is estimated by the flow). This is a simplifying assumption, since it is possible that the picked point will shift slightly after it is released by the gripper; our method does not take into account such small movements.
Using this assumption, to compute the place points $q_1, q_2$, we query the flow $f$ at each pick point $p_1, p_2$ to estimate the delta between the pick point location in the observation image and the corresponding location of the pick points in the goal image. We use these predicted correspondences as the place points:
$q_i = f(p_i) + p_i$, for each arm $i$.

Action predictions estimated by our approach can produce nearly overlapping pick and place points, indicating that arm 1 and arm 2 should perform identical actions. 
We observe this behavior from PickNet when the goal is best achieved with a single-arm action, rather than a bimanual one.
\hl{On a real robot, grippers are likely to collide if grasping points that are too close.}
Therefore, to switch between executing a single-arm or bimanual action, we compute the L2 pixel distance between pick points $d_\text{pick} = \| p_1 - p_2 \|_2$ and place points $d_\text{place} = \| q_1 - q_2 \|_2$. 
We use a single-arm action when either distance is smaller than a threshold $\alpha$, which we set to 30 for all experiments.

\subsection{Implementation Details}
\label{sec:implementation}
We use SoftGym~\cite{Lin-2020-127232}, an environment for cloth manipulation built on the particle-based simulator Nvidia Flex, to collect training datasets. The simulator models cloth as particles connected by springs.
We use pickers that simulate a grasping action by binding to the nearest cloth particle within a threshold to execute pick and place actions in SoftGym.
We collect data by taking random actions, biased towards grasping corners of the cloth. 
We demonstrate that we are able to train our method in SoftGym and then transfer the policy to the real world.
Details on the data collection, as well as the network architecture and training details, can be found in \hlc{\mbox{\appsect{sec:ffn-implementation-details}}}.

\section{Experiments}
\label{sec:experiments}

\subsection{Simulation Experiments}
\label{sec:simulation experiments}

\textbf{Experiment Setup.} We evaluate FabricFlowNet (FFN) and compare to state-of-the-art baselines in the SoftGym~\cite{Lin-2020-127232} simulator; real-world evaluations are below in~\sect{sec:real_experiments}.
\hl{Our experiments focus on folding tasks, and we assume that a cloth smoothing method (\eg \mbox{\cite{ha2021flingbot, seita_fabrics_2020}}) is used to flatten the cloth before folding is executed.}
Our error metric is the average particle position error between the achieved and goal cloth configuration. 
We evaluate on two sets of goals: 40 \textit{one-step} goals that can be achieved with a single fold action, and 6 \textit{multi-step} goals that require multiple folding actions. 
The multi-step goals each consist of a sequence of sub-goal images, with the next sub-goal presented after each action. 
This protocol follows from our problem formulation in~\sect{sec:problemdef}, and is similar to the protocol in Nair~\etal\cite{nair2017combining}.
Our goals include test goals from Ganapathi~\etal\cite{ganapathi2020learning} and Lee~\etal\cite{lee2020learning} that are achievable with one arm, as well as additional goals more suitable for two-arm actions (see \appfig{fig:sim_goals_results} for the full set of goals). 

We compare our method to  
Fabric-VSF~\cite{Hoque-RSS-20}, which learns a visual dynamics model and uses CEM to plan using the model. We only use Fabric-VSF with RGB-D input, as depth-only input performs poorly for folding tasks~\cite{Hoque-RSS-20}. FabricFlowNet only uses depth and does not rely on RGB, which enables our method to transfer easily to the real world without extensive domain randomization. 
We also compare to Lee~\etal\cite{lee2020learning}, a model-free approach. 
We extend the the original single-arm method to a dual-arm variant and compare against both. 
For both our method and the baselines, we only allow each method to perform one pick-and-place action for each subgoal (e.g. one pick and place action for each single-step goals).
Additional baseline details can be found in the Appendix.

\subsubsection{Simulation Results}

\tbl{table:sim_results} contains our simulation results for all methods. We report average particle distance error (in mm) for one-step goals only, multi-step goals only, and over both one-step and multi-step goals.
\hl{Our results show that FFN achieves the lowest error over all goals and has the fastest inference time.} 

\begin{table*}[ht!]
    \centering
    \caption{\hl{Mean Particle Distance Error (mm) and Inference Time (sec) on Cloth Folding Goals}}
    \label{table:sim_results}
    \normalsize
    \begin{tabular}{l d{2.2} d{2.2} d{2.4} | D{.}{.}{4,4}}
      \toprule
        Method 
            & \multicolumn{1}{c}{One Step (n=40)} 
            & \multicolumn{1}{c}{Multi Step (n=6)} 
            & \multicolumn{1}{c|}{All (n=46)}
            & \multicolumn{1}{c}{Inference Time} \\
        \midrule
        Lee~\textit{et al.}, 1-Arm~\cite{lee2020learning}
            & 16.18,08.38 & 26.20,16.31 & 17.49,10.10 & \sim0.04\\
        Lee~\textit{et al.}, 2-Arm
            & 36.62,14.51 & 47.71,21.95 & 38.07,15.82 & \sim0.04\\
        Fabric-VSF~\cite{Hoque-RSS-20}
            & 6.31,06.55 & \textbf{21.33},\textbf{11.20} & 8.27,08.90 & \sim420\\
        FabricFlowNet (Ours) 
            & \textbf{4.46},\textbf{02.62} & 25.04,22.88 & \textbf{7.14},\textbf{11.06} & \mathbf{\sim0}.\mathbf{007}\\
      \bottomrule
    \end{tabular}
    \vspace{-1em}
\end{table*}

% \textbf{\textcolor{red}{
We also investigate whether using flow as a goal recognizer improves performance. 
When an observation closely matches the goal, the flow for all points is close to zero. 
We leverage this fact by evaluating FFN with ``iterative refinement'': we allow the policy to take multiple actions per subgoal to try to further minimize the flow between the observation and subgoal. When the average flow between observation and current subgoal reaches a minimum threshold, the policy moves forward to the next subgoal. FFN with iterative refinement achieves a mean error of 6.62 over all goals vs. 7.14 without refinement. Additional details on iterative refinement can be found in the Appendix, along with additional results from baseline variants, crumpled initial configurations, and end-to-end training.
% }} 

% \hl{Images of goals, as well as additional results on baseline variants, \hlc{iterative refinement,} crumpled initial configurations,}\hl{ and end-to-end training} can be found in the Appendix.

\subsubsection{Ablations}
\label{sec:ablations}

We run series of ablations to evaluate the importance of the components of our system; results \hl{averaged over all 46 goals} are in Table~\mbox{\ref{table:ablations}}. \hlc{Additional details and results are in \mbox{\appsect{sec:app_ablations}}}. 
Our ablations are designed to answer the following questions:

\noindent \textbf{What is the benefit of using flow as input?} We modify PickNet to receive depth images of the observation and goal as input to the network (``NoFlowIn''), as is commonly done in previous work on goal-conditioned RL~\cite{lee2020learning, seita2020learning}.
In this ablation, the PickNet needs to reason about both the relationship between the observation and the goal, as well as the action.
\hlc{In contrast, our method uses the flow network to compare the observation and goal; the picknet separately reasons about the action.}
\textbf{What is the benefit of using flow to choose the place point?} In this ablation, we train a network to predict the place points directly (``NoFlowPlace"). This is in contrast to our approach where we use the flow field, evaluated at the pick point $f(p_i)$, to compute the place point $q_i$ for arm $i$. Our approach leads to a 32.4\% improvement, showing the benefit to using flow as an action representation.\\
\hlc{\mbox{\textbf{What is the performance with no flow?}}} We combine the above two ablations and remove flow entirely, (``NoFlow"; ours has 60.4\% improvement). The above ablations all indicate the strong benefit of using flow as both an input and action representation for cloth manipulation. \\
\textbf{What is the benefit of biasing the data collection to grasp corners?} Our method uses prior knowledge about cloth folding tasks to bias the training data and pick at corners of the cloth. In this ablation, we choose pick points randomly (``NoCornerBias", ours has 35.5\% better performance). \\
\hlc{\mbox{\textbf{What is the performance with a simpler architecture?}}} We also compare our architecture for PickNet (\sect{sec:picknet}) to a simpler architecture that takes as input the flow image $I_f$ and outputs a two heatmaps, one for each pick point (``NoSplitPickNet"; ours has 2.1\% better performance). \\
\textbf{Does the loss formulation in Eq.~\ref{eq:lpick} improve performance?} We compare our method to an ablation where the first ground-truth heatmap is used to supervise PickNet1 and similarly for the second, i.e. $\mathcal{L}_\text{Pick} = l_\text{BCE}(H_1, H_2, H_1^*, H_2^*)$. ("NoMinLoss"; ours has similar performance).

\begin{table*}[ht!]
    \caption{\hl{Mean Particle Distance Error (mm) for Ablations over All Goals (n=46)}}
    \label{table:ablations}
    \small
    \begin{tabular}{c c c c c c | c}
      \toprule
        NoFlowIn
        & NoFlowPlace 
        & NoFlow
        & NoCornerBias 
        & NoSplitPickNet
        & NoMinLoss
        & \hl{FFN (Ours)}
        \\
        \midrule
        % \emph{Error} 
        % & 
        9.37
        & 10.56 
        & 18.02 
        & 11.07 
        & 7.29
        & 7.15
        & $\mathbf{7.14}$
        \\
      \bottomrule
    \end{tabular}
    \vspace{-1em}
\end{table*}

\subsection{Real World Experiments}
\label{sec:real_experiments}

We evaluate FabricFlowNet in the real world and demonstrate that our approach successfully manipulates cloth on a real robot system. 

\textbf{Experiment Setup.} Our robot system consists of two 7-DOF Franka Emika Panda arms and a single wrist-mounted Intel RealSense D435 sensor (See \fig{fig:overview}). 
We plan pick and place trajectories using MoveIt!~\cite{chitta2012moveit}. 
We evaluate on a 30x30 cm towel, using \hlc{6} single-step and \hlc{5} multi-step goals (see \fig{fig:real_results}) that form a representative subset of our simulation test goals. 

To transfer
from simulation to the real world, 
we align the depth between real and simulated images by subtracting the difference between the average depth of the real support surface (i.e. the table) and the simulated surface. 
\hl{We mask the cloth by color-thresholding the background; see Appendix for details}.
We found that these simple techniques were sufficient to transfer the method trained entirely in simulation to the real world, because we use only depth images as input. Simulated depth images match reasonably well to real depth images, unlike RGB images. 

\subsubsection{Real World Results}

\begin{figure}[ht!]
    \centering
    \begin{subfigure}[t]{0.54\textwidth}
        % https://docs.google.com/drawings/d/1DwdPcOsruEywUd3Jqsvoxd2ifrO4hiQTXI0n2n60W0A/edit?usp=sharing
        % \includegraphics[width=\textwidth,natwidth=3149,natheight=1052]{figures/real_cloth/towel_single_mainpaper_viz.png}
        \includegraphics[width=\textwidth]{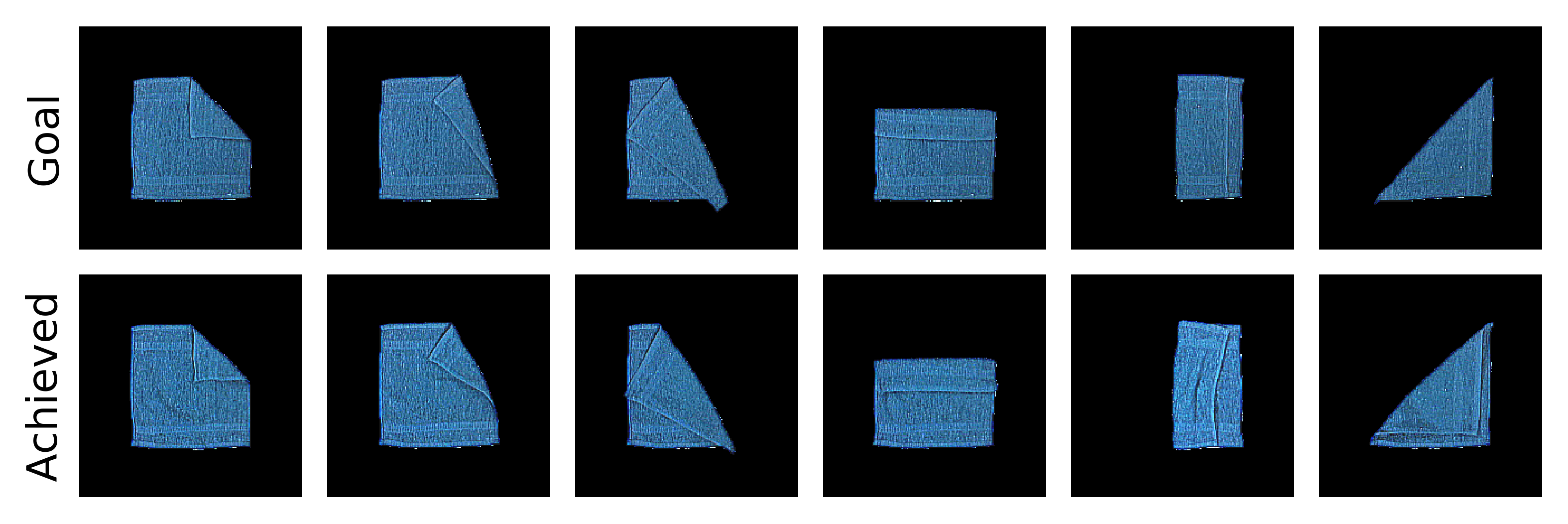}
        \caption{\hlc{One-step Square Cloth}}
    \end{subfigure}
    \begin{subfigure}[t]{0.45\textwidth}
        % https://docs.google.com/drawings/d/1Uq_PmFa_dzP6eKeXf2wKRJHcmsDMQ6DiaGqApSaaslI/edit?usp=sharing
        % \includegraphics[width=\textwidth,natwidth=1252,natheight=3149]{figures/real_cloth/towel_multi_mainpaper_viz.png}
        \includegraphics[width=\textwidth]{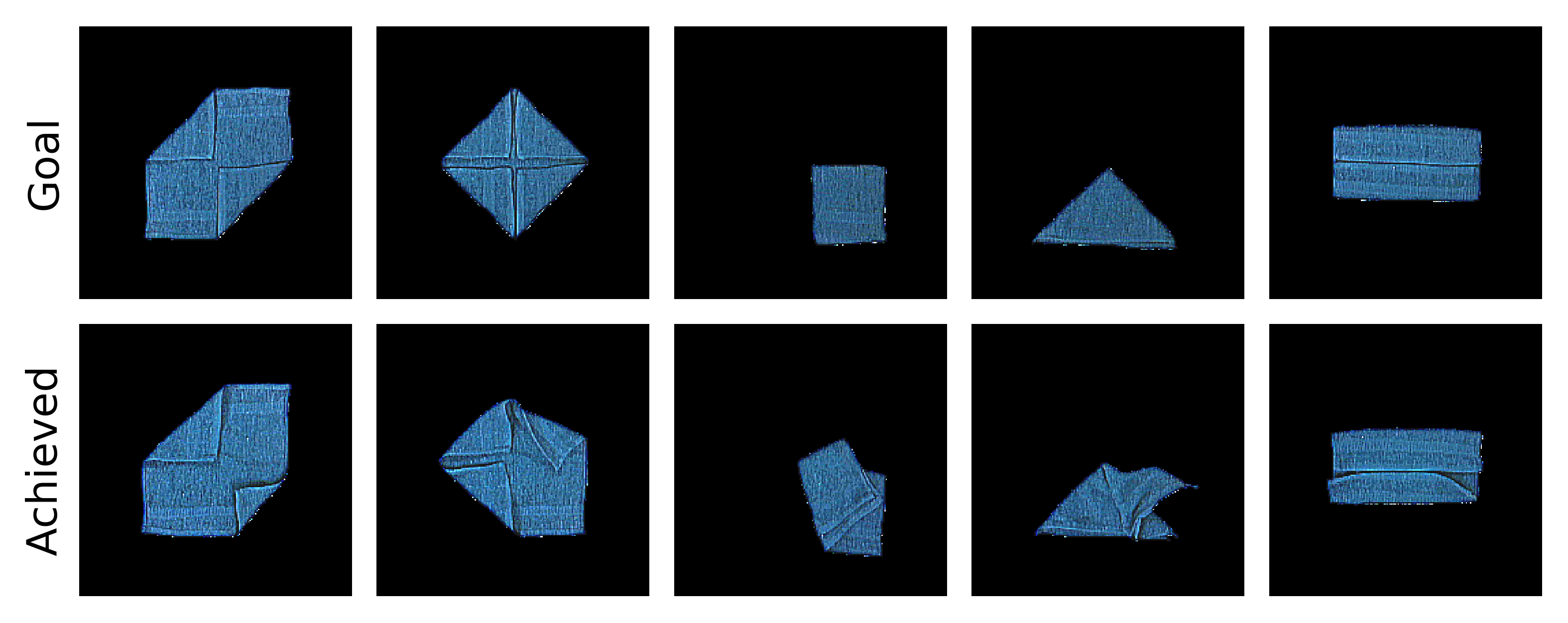}
        \caption{\hlc{Multi-step Square Cloth}}
    \end{subfigure}
    \caption{Qualitative results for FFN on real world experiments. FFN only takes depth images as input, allowing it to easily transfer to cloth of different colors.
    } 
    \label{fig:real_results}
    % \vspace{-1em}
\end{figure}

\fig{fig:real_results} provides qualitative real world results, showing that we successfully achieve many of the goals. 
Our website (link in abstract) contains videos of these trials.

We compare FabricFlowNet to the NoFlow ablation from \sect{sec:ablations}. Both methods used the same sim-to-real techniques described in the previous section.
While we do not have access to the true cloth position error in the real world, Intersection-over-Union (IoU) on the achieved cloth masks serves as a reasonable proxy metric~\cite{lee2020learning}. FFN achieves \hlc{0.80 mean IoU over 3 trials for the square cloth, compared to 0.53 for NoFlow}.
See the Appendix for additional details.

\begin{figure}[ht!]
    \centering
    % \begin{subfigure}[t]{0.42\textwidth}
    \begin{subfigure}[t]{0.54\textwidth}
        \includegraphics[width=\textwidth]{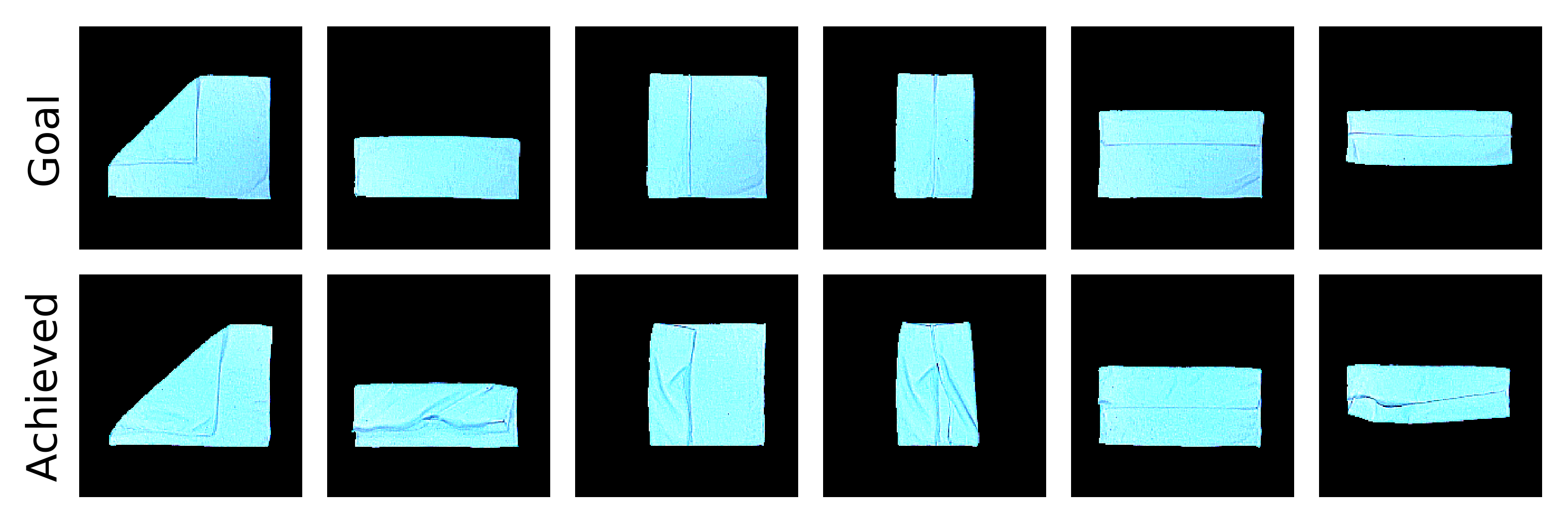}
        \caption{\hlc{Rectangular Cloth}}
    \end{subfigure}
    % \begin{subfigure}[t]{0.52\textwidth}
    \begin{subfigure}[t]{0.45\textwidth}
        \includegraphics[width=\textwidth]{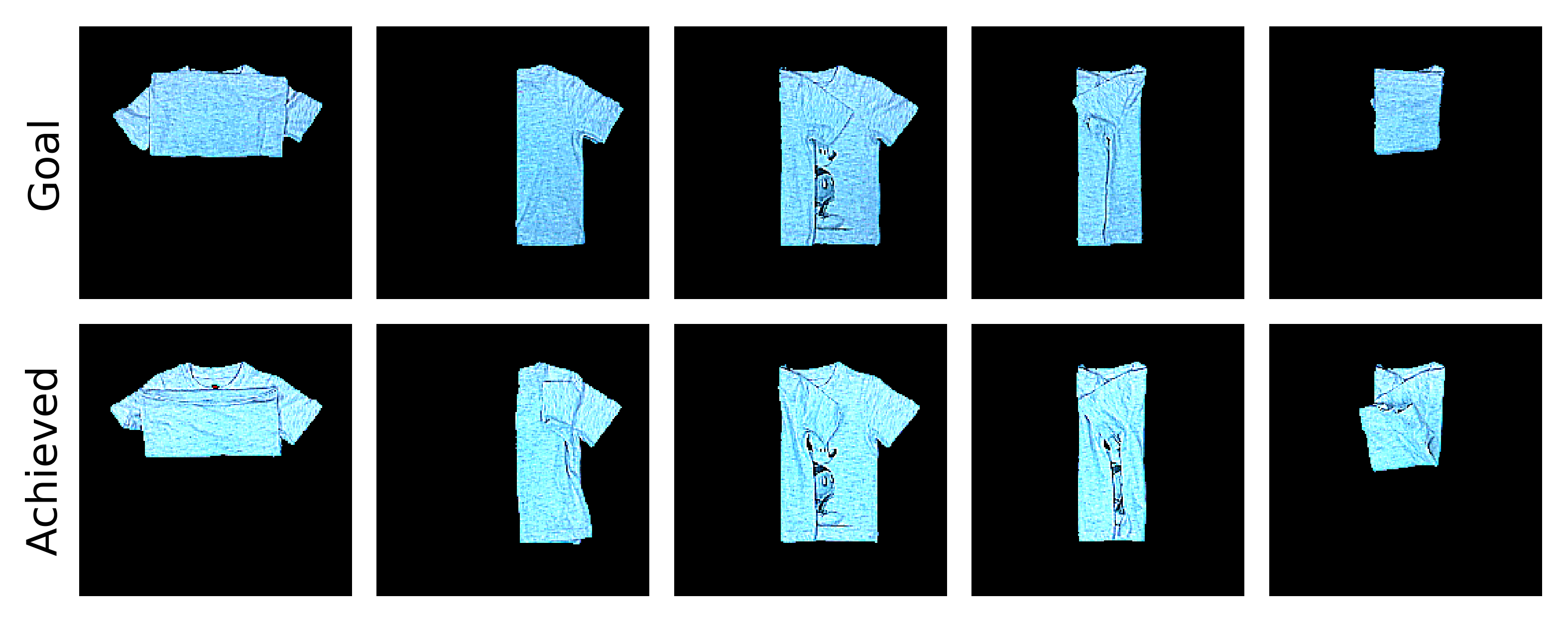}
        \caption{\hlc{Printed T-shirt}}
        \end{subfigure}
        \caption{Generalization to new cloth shapes for FFN trained only on a square cloth in simulation. FFN achieves single and multi-step goals for rectangular fabric and a printed T-shirt.
        } 
    \label{fig:generalization}
    % \vspace{-1em}
\end{figure}

\textbf{Generalization.} In addition to evaluating the folding policy on square cloth for various goal configurations, we also test the generalization of our method to other shapes of cloth. We evaluate the performance of FFN trained only on a square cloth on folding goals for a rectangular cloth as well as a T-shirt. 
These fabrics are also thinner than the square blue towel used in the real world experiments above.
\fig{fig:generalization} shows that FFN trained on a square yellow cloth in simulation is able to generalize to other cloth shapes, textures, and colors (FFN only receives depth images as input).
See Appendix for additional details.
\section{Conclusion} 
\label{sec:conclusion}

In this work we present FabricFlowNet, a method which utilizes flow to learn goal-conditioned fabric folding. We leverage flow to represent the correspondence between observations and goals, and as an action representation. The method is trained entirely using random data in simulation. Our results show that separating the correspondence learning and the policy learning can improve performance on an extensive suite of single- and dual-arm folding goals in simulated and real environments. Our experiments also demonstrate generalization to different fabric shapes, textures, and colors. \hl{Future work on flow-based fabric manipulation could incorporate actions beyond pick and place, such as parameterized trajectories or dynamic actions.}

\clearpage
% The acknowledgments are automatically included only in the final and preprint versions of the paper.
\acknowledgments{This work was supported by 
the National Science Foundation (NSF) Smart and Autonomous Systems Program (IIS-1849154), 
a NSF CAREER Award (IIS-2046491),
LG Electronics, and a NSF Graduate Research Fellowship (DGE-1745016).}

%===============================================================================

% \bibliography{main}  % .bib
\putbib[main]
\end{bibunit}

\clearpage
% \bibliography{supplement/appendix}
% \part*{Part 2}
\begin{bibunit}

\appendix
% \begin{document}
% \section{Appendix}
% \label{app:appendix}
\beginsupplement

% \tableofcontents
\etocsetlevel{appendixplaceholder}{-1}
\etoctoccontentsline*{appendixplaceholder}{APPENDIX}{-1}
\localtableofcontents

\section{Additional Details and Results for FabricFlowNet}

\subsection{FabricFlowNet Implementation Details}
\label{sec:ffn-implementation-details}

\textbf{Data Collection}.
We collect data in SoftGym by taking random pick and place actions on the cloth. 
The random actions are biased to pick corners of the cloth mask (detected using Harris corner detection \cite{Harris1988ACC}) 45\% of the time, and ``true" corners of the square cloth 45\% of the time. If the true corners are occluded then Harris corners are used instead. For the remaining 10\%, the pick actions are uniformly sampled over the visible cloth mask.
After the pickers grasp the cloth, they lift to a fixed height of 7.5 cm. 

\begin{figure}[ht!]
    \centering
    \includegraphics[width=0.4\textwidth]{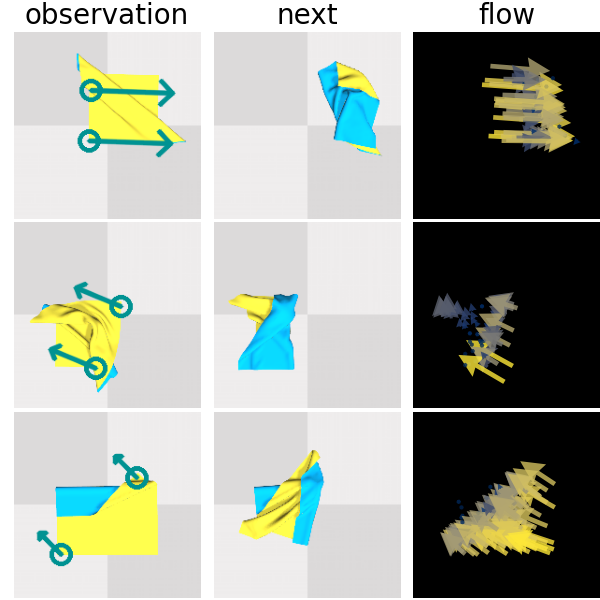}
    \caption{Training data for FFN.} 
    \label{fig:data}
\end{figure}

We constrain the place points of the action so that both place points are offset in the same direction and distance from their respective pick points. 
The direction is orthogonal to the segment connecting the two pick points, and points towards the center of the image, so the cloth does not move out of the frame (similar to Lee~\etal\cite{lee2020learning}).
The distance between the pick point and the place point along this direction is uniformly sampled between \hl{$[25, 150]$} px.
The distance is truncated if it exceeds a margin of 20 px from the image edge, again to prevent moving the cloth out of the frame.
While these heuristics may seem to overly constrain the data we collect, we observe that our data still contains highly diverse cloth configurations, as shown in \fig{fig:data}.

For each sample, we save the initial depth observation image, the dual-arm pick and place pixel locations of the action, the next depth observation resulting from the executed action, and the cloth particle positions of both observations (See \fig{fig:data}).
The camera for capturing depth observations is fixed at 65 cm above the support surface.
We mask the depth observations to only include the cloth by setting all background pixels to zero. 
The dataset for training both the flow and pick networks consists of 20k samples from \hlc{4k}
% 10k 
episodes, where each episode consists of \hlc{five} dual-arm pick and place actions.

\textbf{Flow Network Training.}
We use FlowNet~\cite{dosovitskiy2015flownet} as our flow network architecture.
The input to FlowNet is the initial and next depth image from a sample in our dataset, stacked channel-wise.
The ground truth flow for supervising FlowNet comes from the cloth particles used by the simulator to model the cloth's dynamics: we collect the cloth particle positions for each observation in our dataset and correspond them across observations to get flow vectors (See~\fig{fig:data}). 
The ground truth flow is sparse because the cloth particles are sparse, so we train FlowNet using a masked loss that only includes pixels with corresponding ground truth flow.
Similar to Lee~\etal\cite{lee2020learning}, we apply spatial augmentation of uniform random translation (up to 5 px) and rotation (up to 5 degrees) to augment the training data. 
We train the network using the dataset of 20k random actions described above.
We use the Adam~\cite{kingma2014adam} optimizer, learning rate $1\text{e-}4$, weight decay $1\text{e-}4$, and batch size 8.

\textbf{PickNet Network Training.}
PickNet1 and PickNet2 are fully-convolutional network architectures based on Lee~\etal\cite{lee2020learning}, with 4 convolutional layers in the encoder, each with 32 filters of size 5. The first three layers of the encoder have stride 2 and the last one has stride 1. The decoder consists of 2 interleaved convolutional layers and bilinear upsampling layers. 

The input to the PickNet1 is a 200 $\times$ 200 flow image. PickNet2 receives the first pick point location (the argmax of the Picknet1 output, as described in the main text) as an additional input, represented as a 2D Gaussian $\mathcal{N}(p_1, \sigma)$ (where $\sigma=5$). 
Similar to Nair~\etal\cite{nair2017combining}, the output of both networks is a 20 x 20 spatial grid.
If the pick points predicted by PickNet are not on the cloth mask, we project them to the closest pixel on the mask using an inverse distance transform.
In practice, we find that the predictions are usually either on the cloth mask or very close to the mask.
To train PickNet1 and PickNet2, we use the same dataset of 20k random actions described above. We use the Adam~\cite{kingma2014adam} optimizer, learning rate $1\text{e-}4$, and batch size 10.

\begin{figure}[ht!]
    \begin{subfigure}[t]{0.49\textwidth}
        \centering
        \includegraphics[width=\textwidth]{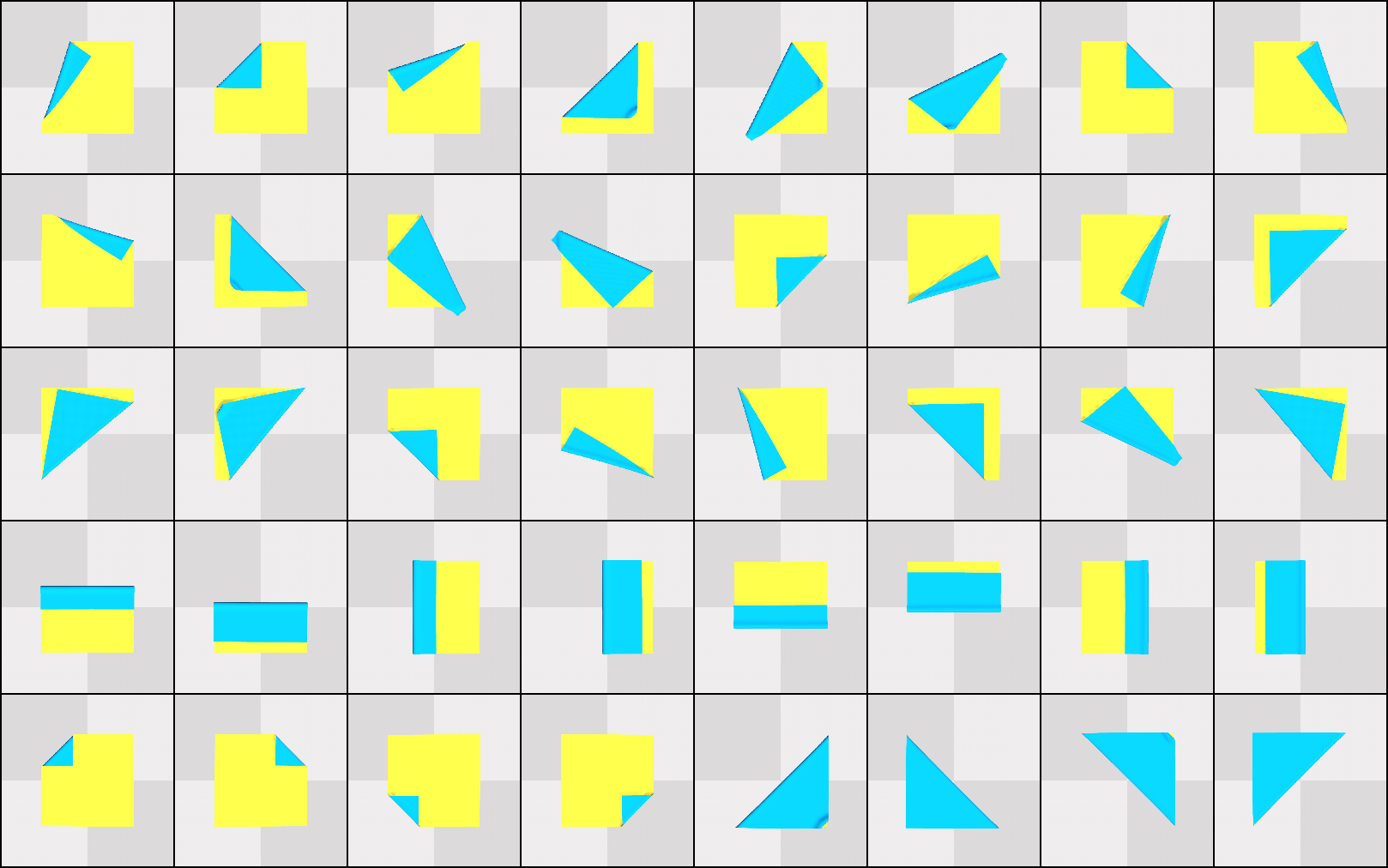}
        \caption{One-step goals} 
        \label{fig:sim_goals}
    \end{subfigure}
    \hfill %%
    \vspace{0.5em}
        % \hspace*{10px}
    \begin{subfigure}[t]{0.49\textwidth}
        \centering
        \includegraphics[width=\textwidth]{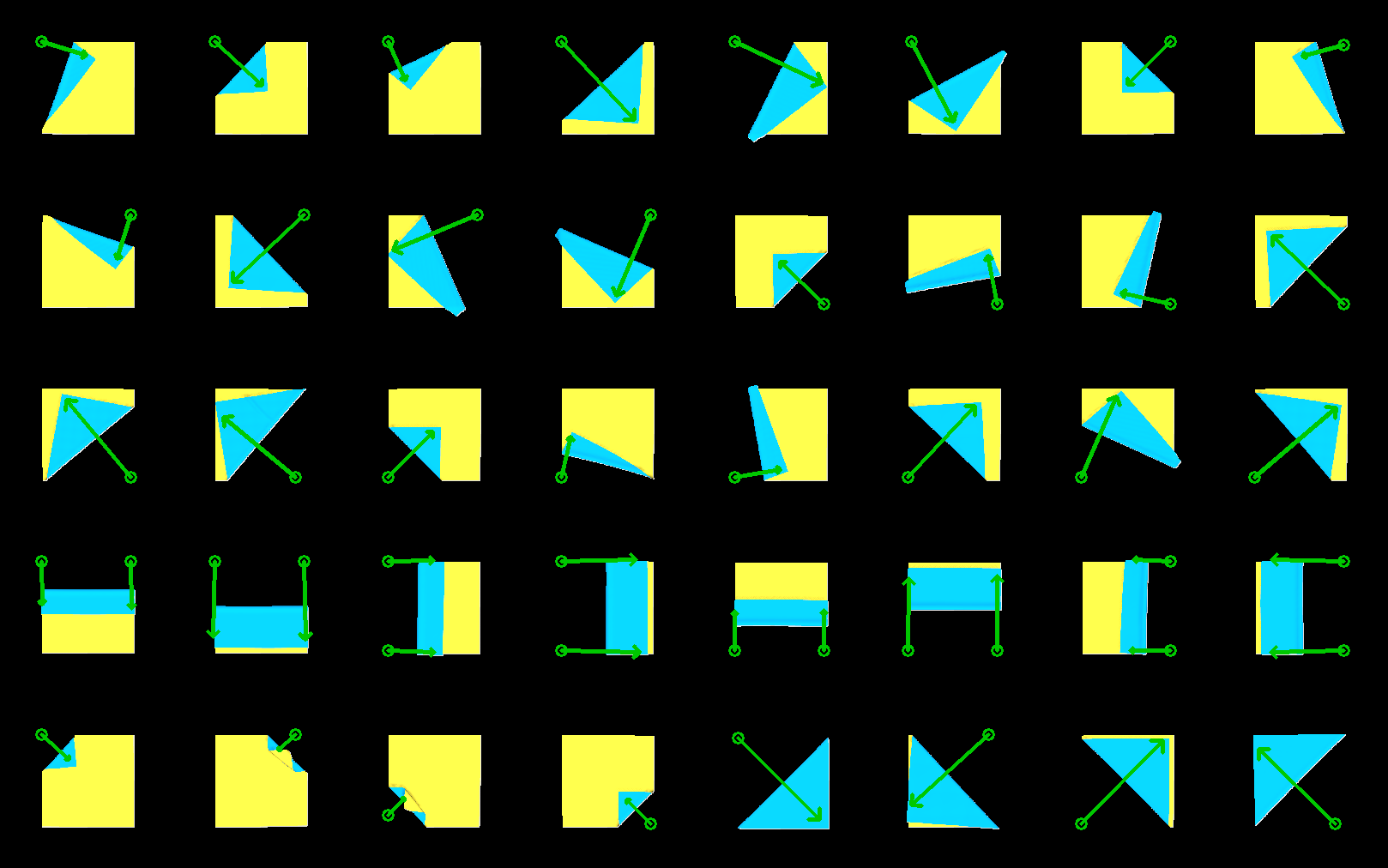}
        \caption{One-step FFN performance}
        \label{fig:sim_results}
    \end{subfigure}
    \hfill
    \newline
    \begin{subfigure}[t]{0.49\textwidth}
    \centering
    \includegraphics[width=\textwidth]{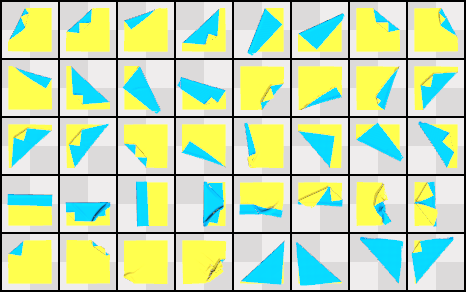}
    \caption{One-step Fabric-VSF performance}
    \label{fig:sim_vsf}
    \end{subfigure}
    \hspace{1px}
    \begin{subfigure}[t]{0.49\textwidth}
    \centering
    \includegraphics[width=\textwidth]{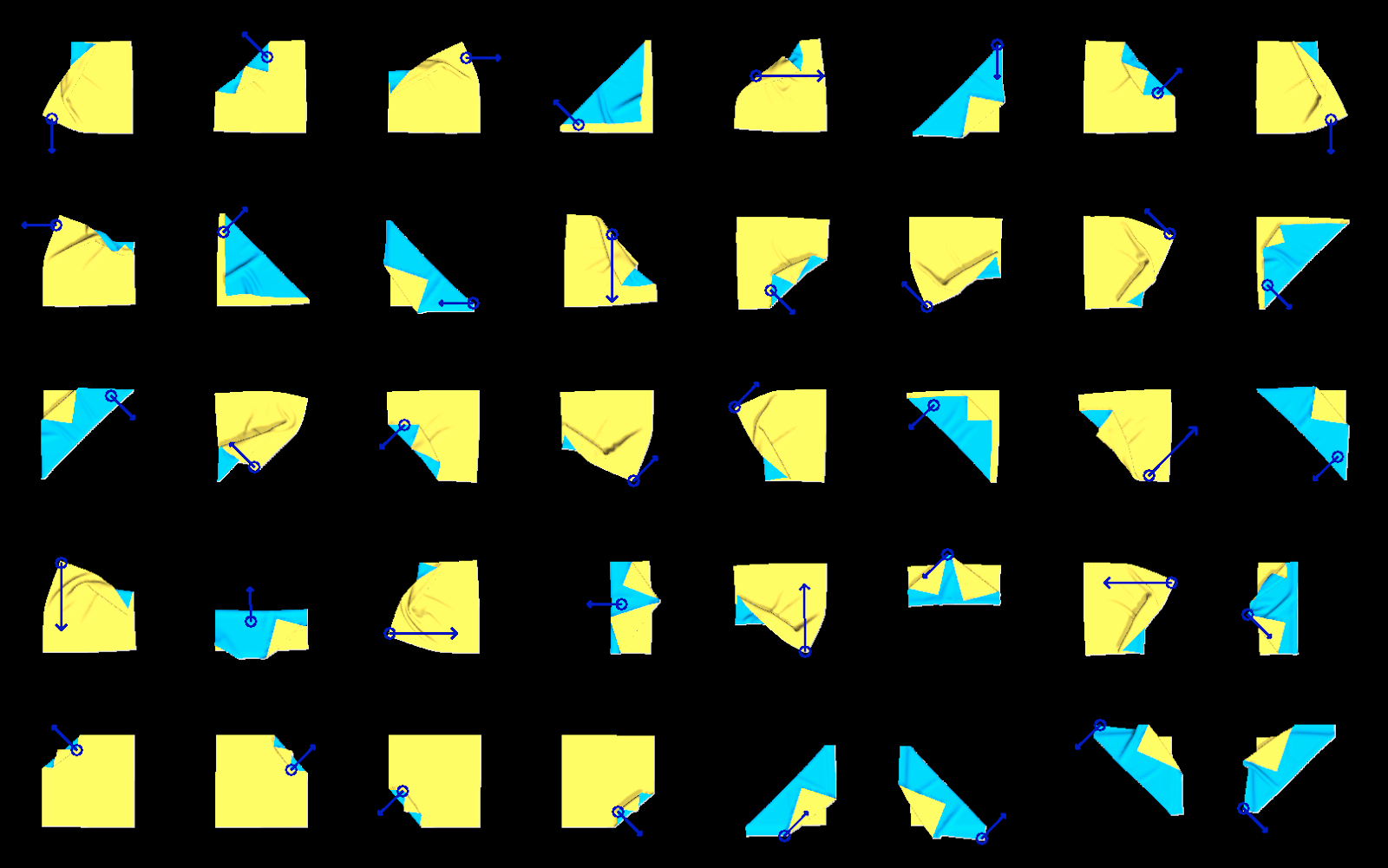}
    \caption{One-step Lee et al. performance}
    \label{fig:sim_lee}
    \end{subfigure}
    \hfill %%
    \vspace{0.5em}
    \newline
     \begin{subfigure}[t]{0.245\textwidth}
        \centering
        \includegraphics[width=\textwidth]{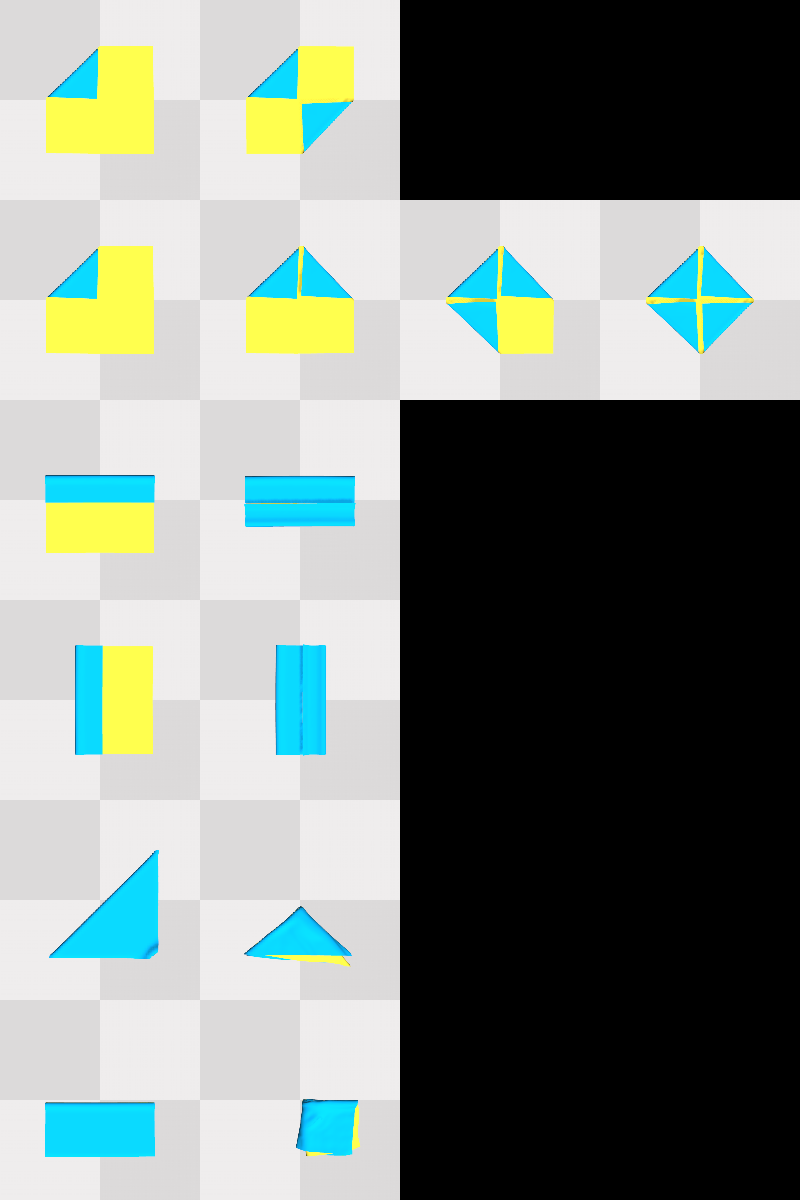}
        \caption{Multi-step goals}
        \label{fig:sim_goals_ms}
    \end{subfigure}
    \begin{subfigure}[t]{0.245\textwidth}
        \centering
        \includegraphics[width=0.99\textwidth]{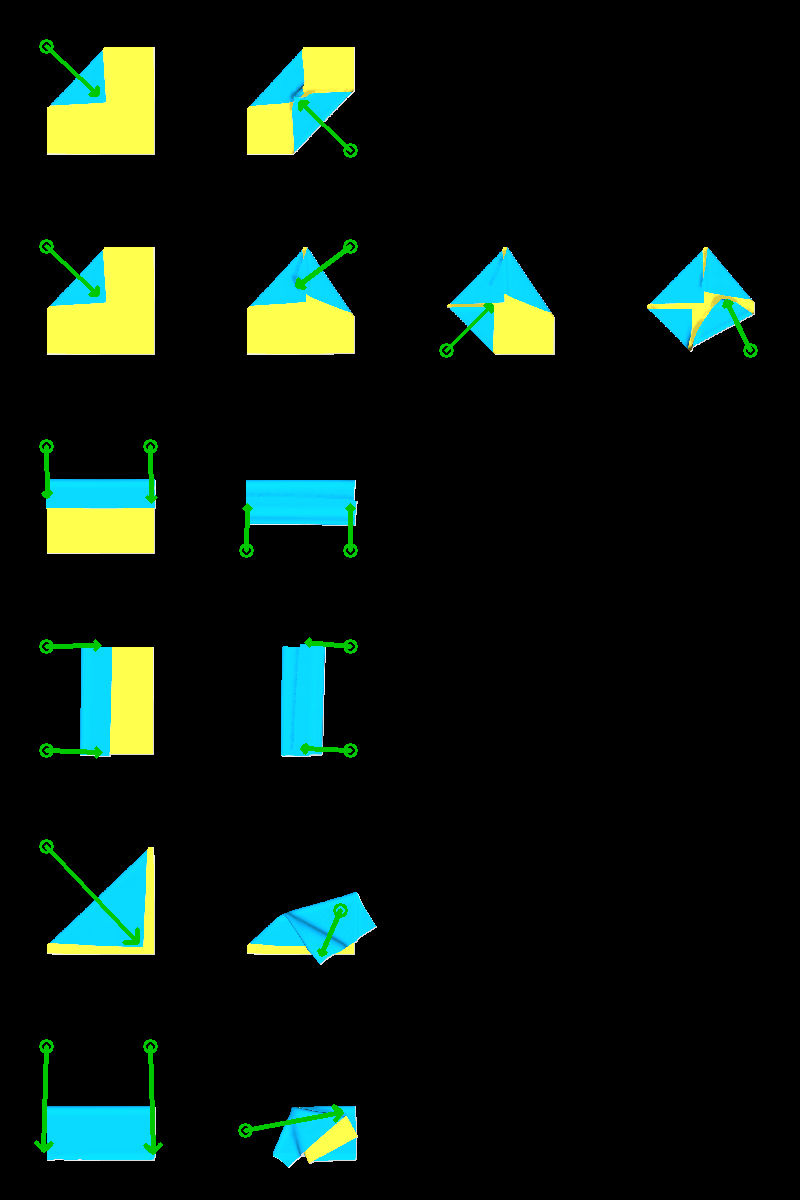}
        \caption{Multi-step FFN performance}
        \label{fig:sim_ms_results}
    \end{subfigure}
    \begin{subfigure}[t]{0.245\textwidth}
        \centering
        \includegraphics[width=0.99\textwidth]{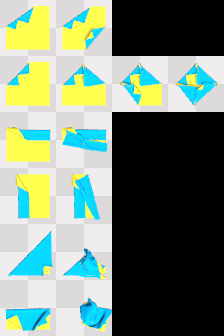}
        \caption{Multi-step Fabric-VSF performance}
        \label{fig:sim_vsf_ms_results}
    \end{subfigure}
    \begin{subfigure}[t]{0.245\textwidth}
        \centering
        \includegraphics[width=0.99\textwidth]{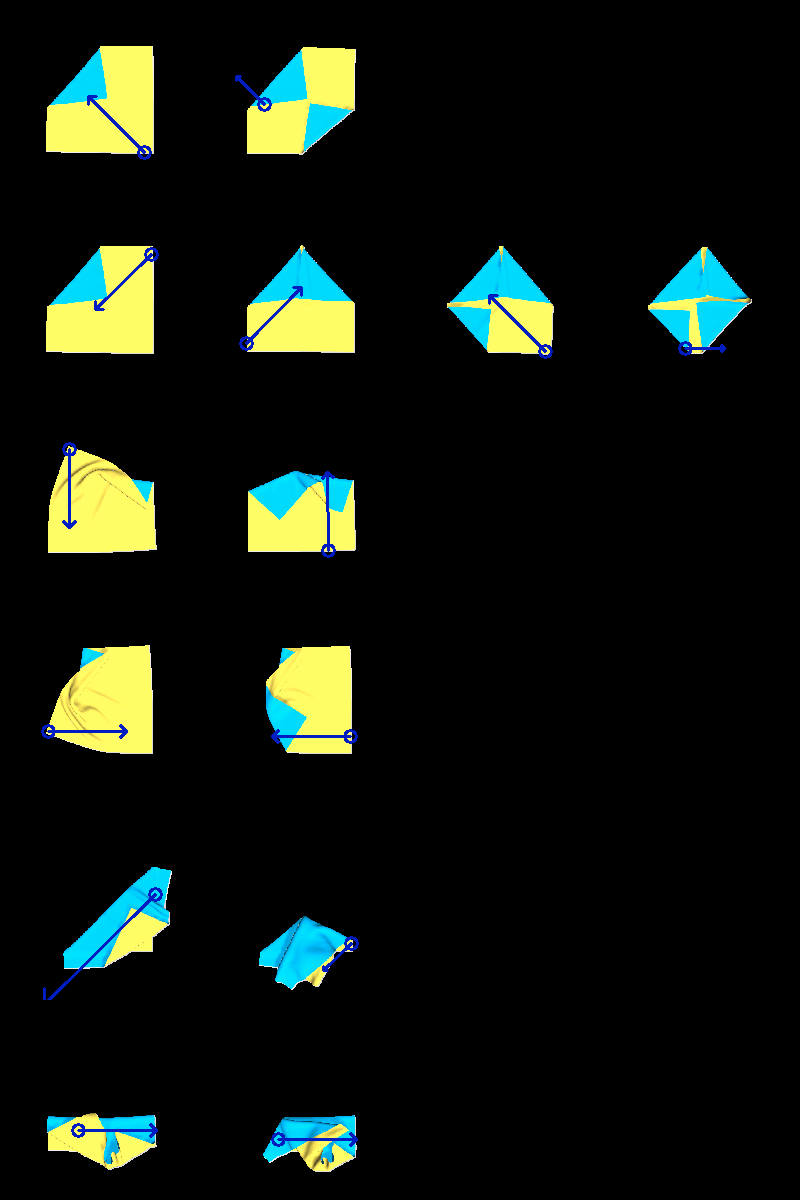}
        \caption{Multi-step Lee et al. performance}
        \label{fig:sim_lee_ms_results}
    \end{subfigure}
    
    \caption{Goal configurations, achieved configurations, and training data in simulation. Arrows indicate the executed action. Fabric-VSF uses a lower camera height than FFN (45 cm vs. 65 cm), thus the cloth looks slightly larger.}
    \label{fig:sim_goals_results}
\end{figure}

\subsection{Additional Simulation Results for FabricFlowNet}

\fig{fig:sim_results} and \fig{fig:sim_ms_results} show the cloth configurations achieved by FabricFlowNet for each of the one-step goals (\fig{fig:sim_goals}) and multi-step goals (\fig{fig:sim_goals_ms}). 
\fig{fig:data} provides examples of the data used to train FFN. \hl{Our policy is deterministic and the simulation is near-deterministic, so we only need 1 trial for our simulation experiments (unlike our real world experiments which use 3 trials).}

\subsection{Additional Real World Details and Results for FabricFlowNet}
\label{sec:app_real_ffn}

\hl{\textbf{Cloth Masking.} In simulation, we can obtain a perfect cloth mask. In the real world, we first obtain a background mask of the table using color-based HSV thresholding, which we can determine before the cloth is placed on the table. We then use the inverse of this background mask to obtain a mask of the cloth. Note that while we use background color of the table for cloth masking, the network itself only takes depth input, allowing the network to be robust to colors and patterns on the cloth itself.}

\begin{figure}[ht!]
    \centering
    \begin{subfigure}[t]{0.75\textwidth}
        % https://docs.google.com/drawings/d/1DwdPcOsruEywUd3Jqsvoxd2ifrO4hiQTXI0n2n60W0A/edit?usp=sharing
        % \includegraphics[width=\textwidth, trim={0 1em 0 0},clip,natwidth=3149,natheight=1550]{figures/real_cloth/towel_single.png}  
        \includegraphics[width=0.99\textwidth]{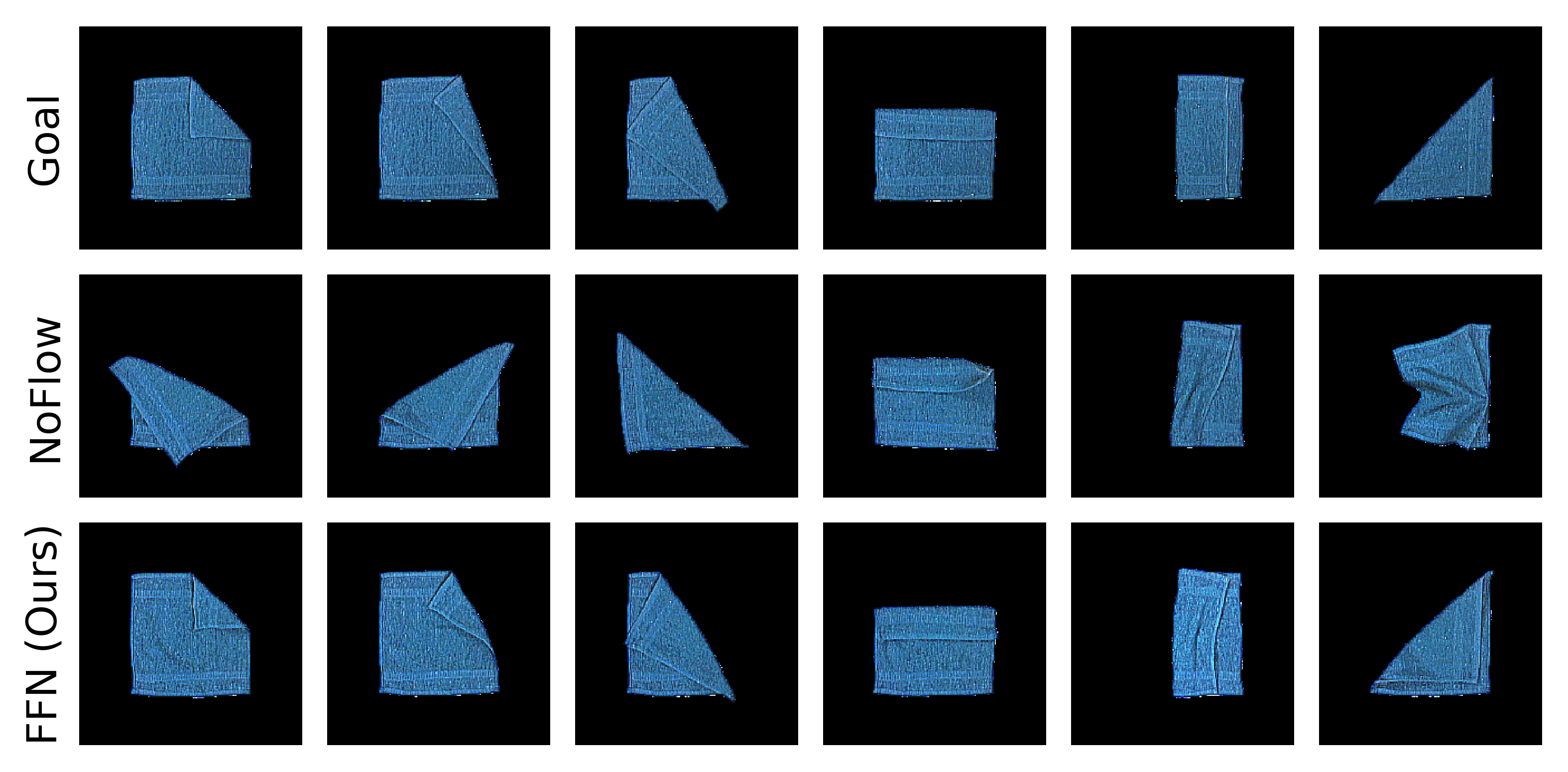}
        \caption*{\hlc{Single-step Square Towel}}
        \vspace{0.5em}
    \end{subfigure}\\
    \begin{subfigure}[t]{0.75\textwidth}
        % https://docs.google.com/drawings/d/1Uq_PmFa_dzP6eKeXf2wKRJHcmsDMQ6DiaGqApSaaslI/edit?usp=sharing
        % \includegraphics[width=\textwidth, trim={0 1em 0 0},clip,natwidth=3149,natheight=1849]{figures/real_cloth/towel_multi.png}
        \includegraphics[width=0.99\textwidth]{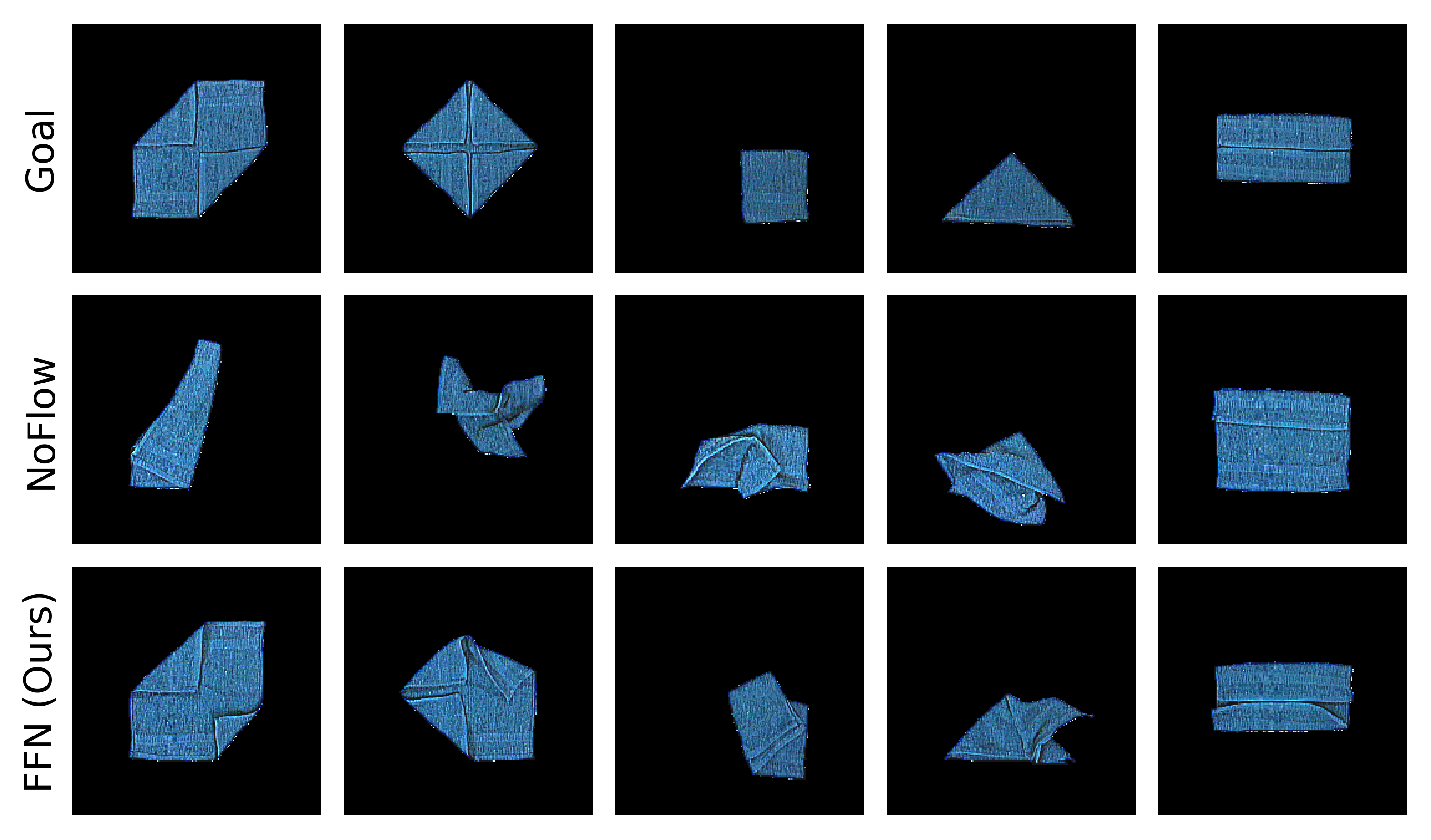}
        \caption*{\hlc{Multi-step Square Towel}}
        \vspace{0.5em}
    \end{subfigure}
    \begin{subfigure}[t]{0.567\textwidth}
        \includegraphics[width=0.99\textwidth]{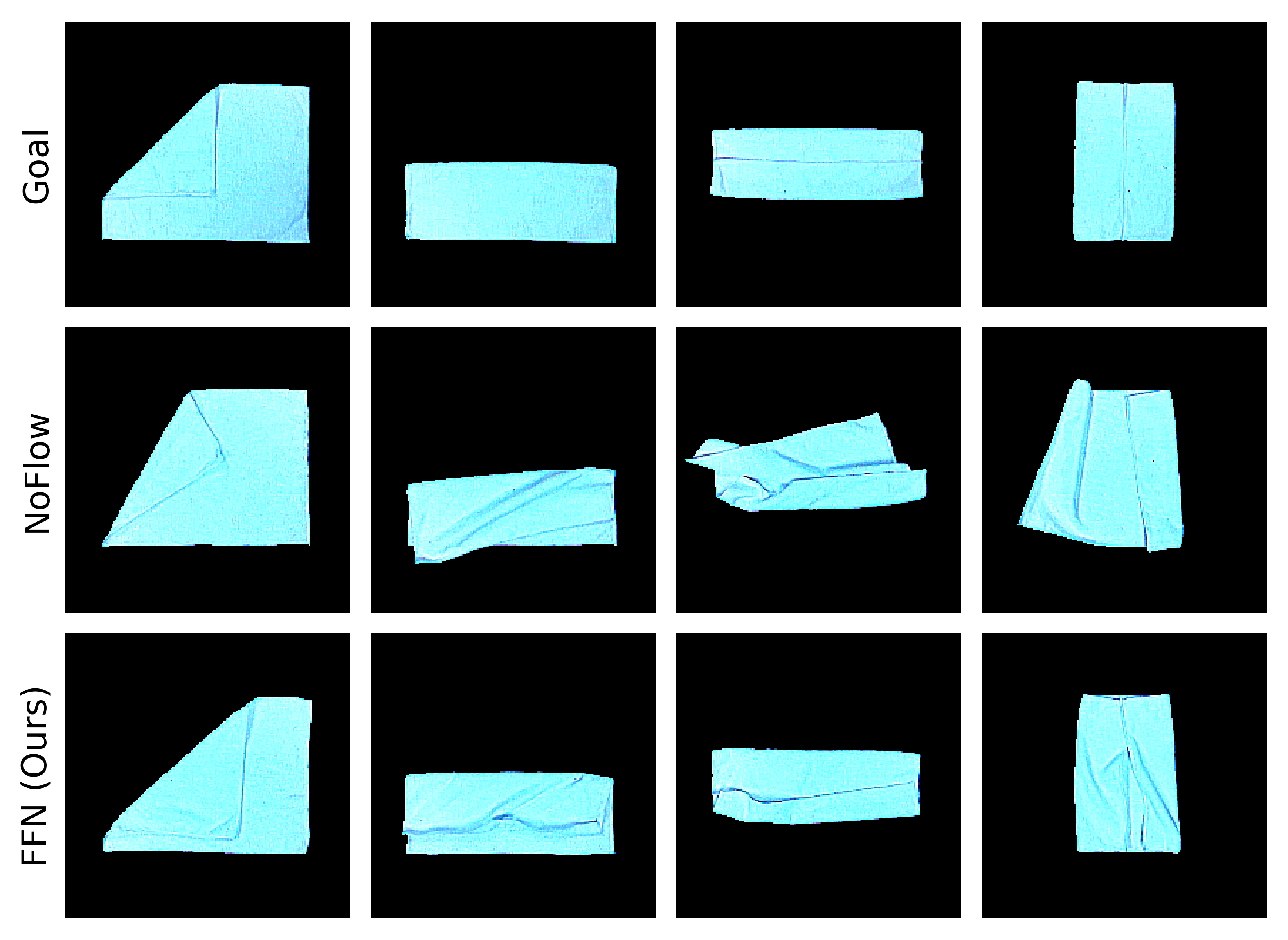}
        \caption*{\hlc{Rectangular Cloth}}
    \end{subfigure}
    \begin{subfigure}[t]{0.422\textwidth}
        \includegraphics[width=\textwidth, trim={0 1em 0 0},clip]{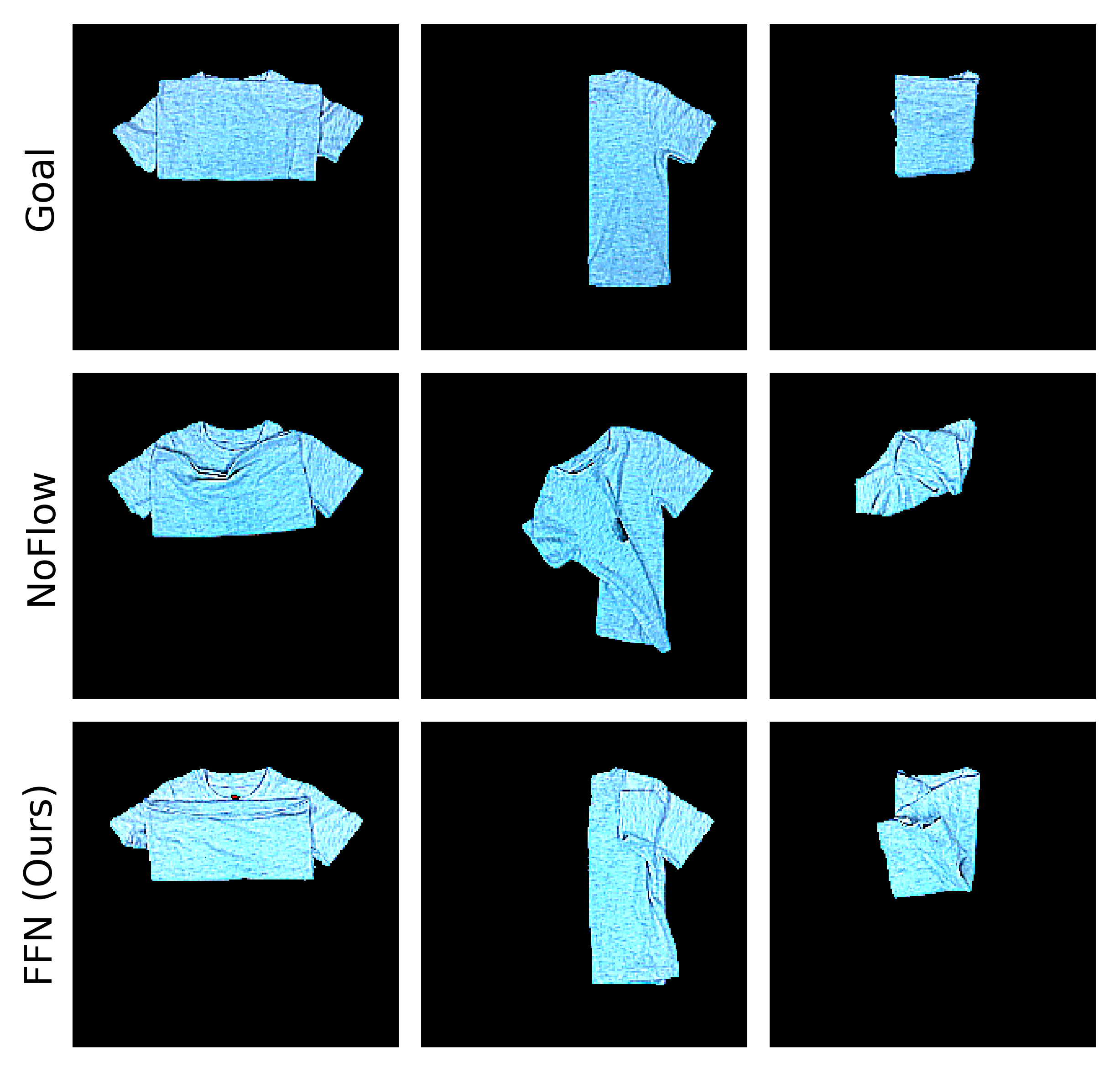}
        \caption*{\hlc{T-shirt}}
    \end{subfigure}
    \caption{\hlc{Qualitative performance of FFN and NoFlow on real cloth. The trial corresponding to the best achieved IOU is shown for each example. For multi-step goals, only the final goal is shown.} FFN only takes depth images as input, allowing it to easily transfer to cloths of different colors. \hlc{Contrast and brightness have been adjusted to enhance visibility.}} 
    \label{fig:real_results_full}
\end{figure}

\textbf{Results on Real Cloth Folding}.
\tbl{table:real_results_mIOU} provides mean IOU (mIOU) performance for NoFlow and FFN on real cloth goals.
The NoFlow ablation performs considerably worse compared to FFN on real cloth folding.
Qualitative results and the complete set of real square cloth goals are in \fig{fig:real_results_full}; the complete set of real rectangle and T-shirt goals are in the main text.
\hlc{We found that for FFN, using FlowNet weights from epochs at the start of convergence transferred better to the real world than using weights from epochs long after convergence.}

\begin{table*}[ht!]
    \centering
    \caption{\hlc{mIOU for Folding Square Towel, Rectangular Cloth, and T-shirt}}
    \label{table:real_results_mIOU}
    \normalsize
    \begin{tabular}{l c c c c c}
      \toprule
        Method & 1-Step Sq. $\uparrow$ & Multi-Step Sq. $\uparrow$ & All Sq. $\uparrow$ & Rect. $\uparrow$ & T-shirt $\uparrow$ \\
         & $(n=6)$ & $(n=5)$ & $(n=11)$ & $(n=3)$ & $(n=3)$ \\
        \midrule
        NoFlow & $0.59\pm0.04$ & $0.45\pm0.01$ & $0.53\pm0.02$ & $0.65\pm0.07$ & $0.61\pm0.06$ \\
        FFN (Ours) & $\mathbf{0.89\pm0.01}$ & $\mathbf{0.69\pm0.04}$ & $\mathbf{0.80\pm0.03}$ & $\mathbf{0.81\pm0.04}$ & $\mathbf{0.82\pm0.02}$ \\
        \bottomrule
    \end{tabular}
    \vspace{5pt}
    \caption*{Average of 3 rollouts. \hlc{Higher mIOU scores are better; the max achievable score is 1.0}.}
\end{table*}

\begin{figure}[ht!]
    \centering
    \begin{subfigure}[t]{0.20\textwidth}
        \includegraphics[width=\textwidth, trim={0 1em 0 0},clip]{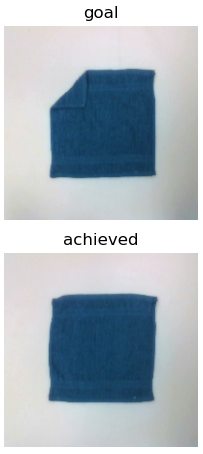}
        \caption{Flopping back}
        \label{fig:fail_flop}
    \end{subfigure}
    \begin{subfigure}[t]{0.20\textwidth}
        \includegraphics[width=\textwidth, trim={0 1em 0 0},clip]{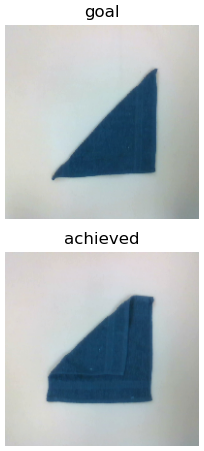}
        \caption{Undershooting}
        \label{fig:fail_flow}
    \end{subfigure}
    \begin{subfigure}[t]{0.20\textwidth}
        \includegraphics[width=\textwidth, trim={0 1em 0 0},clip]{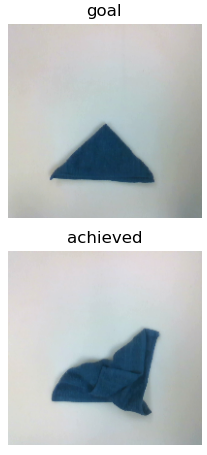}
        \caption{Poor prediction}
        \label{fig:fail_ms}
    \end{subfigure}
    \caption{Examples of failure cases} 
    \label{fig:failure_case}
    \vspace{-0.5em}
\end{figure}

\textbf{Failure Cases}.
This work focused on high level actions with fixed primitives for picking and placing that may not be ideal for all cloth types, sizes, or folds. 
Causes of failures include the grasped portion of the cloth ``flopping back'' against the folding direction, undoing small folding actions or causing unwanted secondary folds (\fig{fig:fail_flop}).
Potential future work is to learn better pick and place primitives.
Another source of failure was over- or under-estimating the fold distance due to slight inaccuracies in the flow prediction (\fig{fig:fail_flow}). 
We also see some failures during multi-step folding; since we provide sub-goals in sequence and allow only one action per sub-goal, the discrepancy between the starting image of the demonstration and the observed image can result in poor predictions (\fig{fig:fail_ms}).
Allowing the policy to take multiple actions to achieve a sub-goal before proceeding may improve performance.
For example, the flow can be recalculated after each action to determine if the observation is sufficiently close to the desired sub-goal configuration before proceeding to the next sub-goal.

\section{Additional Details and Results for Fabric-VSF~\cite{Hoque-RSS-20}}

\subsection{Fabric-VSF~\cite{Hoque-RSS-20} Implementation Details}
The original Fabric-VSF~\cite{Hoque-RSS-20} paper uses single arm actions and a top-down close camera view such that the cloth covers the whole image. To match the camera view, we set the camera height to be 45 cm above the table in our case. The training dataset consists of 7115 trajectories, each with 15 random pick-and-place actions, totaling 106725 data points. Note that this dataset is 5x larger than the 20k samples we train FFN on.
During training, Fabric-VSF takes as input 3 context frames and predicts the next 7 target frames. 

We trained 8 variants of Fabric-VSF. Each variant differs in the following aspects: 1) whether it uses single arm or dual arms; 2) during data collection, whether the pick-and-place actions are randomly sampled, or use the corner biasing sampling strategy as described in Sec.~\ref{sec:ffn-implementation-details}, and 3) whether it uses the original small action size (``Small Action'', bounded to half of the cloth width) or a larger action size (``Large Action'', bounded to the \hl{diagonal length of the cloth}). Other than these three changes, we set all other parameters to be the same as in the original paper. Therefore, the variant with single arm actions, no corner biasing during data collection, and small action size is exactly how Fabric-VSF is trained in the original paper.

After the training, we plan with cross-entropy method (CEM) to find actions for achieving a given goal image. We use the exact same CEM parameters as in the original paper, \ie we run CEM for 10 iterations, each with a population size of 2000 and elite size of 400.

\subsection{Additional Fabric-VSF~\cite{Hoque-RSS-20} Results}
The results for the Fabric-VSF variants are summarized in \tbl{table:fabric_vsf_results}. We note that the variant using single arm actions, corner biasing for data collection, and large action size performs the best out of all variants.
This variant outperforms FFN on overall error and one-step error, but performs slightly worse than FFN on multi-step error (See \fig{fig:sim_vsf} and \fig{fig:sim_vsf_ms_results} for qualitative results). However, we note that Fabric-VSF was trained on 5x more data than FFN. 
Additionally, Fabric-VSF takes much longer to run at inference time, requiring $\sim$7 minutes of CEM iterations to compute a single action compared to $\sim$0.007 seconds for a forward pass through FFN. 
7 minutes of CEM planning time is impractical for real-world folding. We also demonstrate in the following section that FFN generalizes to other cloth shapes better than Fabric-VSF.

Analyzing the performance between different Fabric-VSF variants, for single-arm actions, using large actions instead of small actions always leads to better performance. However, this is not true for the dual arm variants. Interestingly, we find that using dual arms tends to result in worse performance compared with using a single arm. The reason for this could be that during CEM planning, dual-arm variants double the action dimension, which increases complexity for CEM and makes it difficult to find optimal actions.

\begin{table*}[ht!]
    \centering
    \caption{\hl{Mean Particle Distance Error (mm) and Inference Time (sec) for Fabric-VSF Variants}}
    \label{table:fabric_vsf_results}
    \normalsize
    \begin{tabular}{l d{2.2} d{2.2} d{2.2} r}
      \toprule
        Baseline & \multicolumn{1}{c}{1-Step (n=40)} & \multicolumn{1}{c}{Multi-Step (n=6)} & \multicolumn{1}{c}{All (n=46)} & Inf. Time\\
        \midrule
        1-Arm, No CB, Sm. Action  &  12.92,13.00 & 46.05,48.07 &  17.24,23.93 & $\sim$420s \\ 
        1-Arm, No CB, Lg. Action  &  10.13,07.33 & 33.06,12.46 &  13.12,11.25 & $\sim$420s \\ 
        1-Arm, CB, Sm. Action  & 14.09,11.36 & 38.68,27.72  & 17.30,16.76 & $\sim$420s \\ 
        1-Arm, CB, Lg. Action  & 6.30,06.55 & \mathbf{21.33},\mathbf{11.20}
 & 8.27,08.90 & $\sim$420s  \\ 
        2-Arm, No CB, Sm. Action  & 24.60,14.69 &  50.26,27.54  & 27.94,19.00 & $\sim$420s\\ 
        2-Arm, No CB, Lg. Action  &  10.98,05.80 & 40.92,18.06 & 14.89,13.17 & $\sim$420s\\ 
        2-Arm, CB, Sm. Action  & 16.21,13.81 & 36.42,26.51 & 18.84,17.43 & $\sim$420s \\ 
        2-Arm, CB, Lg. Action  & 15.58,10.88  & 54.06,26.68  &  20.60,19.07 & $\sim$420s \\
        \midrule
        FFN (Ours) & \mathbf{4.46},\mathbf{02.62} & 25.04,22.88 & \mathbf{7.14},\mathbf{11.06} & $\mathbf{\sim}$\textbf{0.007s} \\
        \bottomrule
    \end{tabular}
    \vspace{2pt}
    \caption*{\footnotesize CB: Corner Bias \quad Sm. Action: Small Action \quad Lg. Action: Large Action}
    \vspace{-1em}
\end{table*}

\section{Additional Details and Results for Lee~\etal\cite{lee2020learning}}

\subsection{Lee~\etal\cite{lee2020learning} Implementation Details}

Lee~\etal\cite{lee2020learning} learns a fabric folding policy for a discrete action space using a fully convolutional state-action value function, or Q-network. Observation and goal images are stacked channel-wise, then duplicated and transformed to form a batch of $m$ image rotations and $n$ scales to represent different pick and place directions and action lengths. 
The whole batch is input to the Q-network to compute the Q-value of executing an action for each rotation and scale at every point on the image.
The action corresponding to the max Q-value from the outputs is executed. 
The discrete action space of $m$ rotations and $n$ action lengths for Lee~\etal\cite{lee2020learning} enables efficient policy learning, but greatly limits the actions of the learned policy compared to FFN. 

We extend Lee~\etal\cite{lee2020learning} from a single-arm approach to a dual-arm one. To represent two pickers instead of one, we input two pairs of observation and goal images to the Q-network. When rotating and scaling the images to represent different actions, the images are constrained to have the same rotation, but are allowed to be scaled differently. In other words, the dual-arm actions are constrained to execute pick and place actions in the same direction, but can have different pick and place lengths. The Q-network outputs a pair (one for each arm) of Q-value heatmaps for every action in the discrete action space (\ie every rotation and scale). The max Q-value in each of the two heatmaps is averaged, and the heatmap pair with the highest averaged Q-value is selected from the set of all discrete rotations and scales. The picker action corresponding to the argmax of each heatmap is executed.

We train each Lee~\etal variant below using hyperparameters similar to the original paper~\cite{lee2020learning}, training for 25k steps with learning rate $1\text{e-}4$, batch size 10, and evaluating performance on test goals every 500 steps to find the best performing step.

\subsection{Additional Lee~\etal\cite{lee2020learning} Results}

\hlc{We trained variants of Lee\mbox{~\textit{et al.}} to compare single-arm vs. dual-arm performance, depth input vs. RGB input, collecting data with corner bias similar to FFN vs. without bias, and using the original close-up image of the cloth (``Low Cam'') vs. images from further away (``High Cam''). 
All variants were trained with 20k training examples.}
We also provide results for two variants of FFN trained on the same amount of data, one where actions are sampled from the discrete action space (\ie discretized action angles and lengths) in Lee~\etal\cite{lee2020learning} (``Discrete Actions''), and the other where actions are sampled using our continuous action space described in \sect{sec:ffn-implementation-details} (``Cont.\ Actions'').
Lee~\etal\cite{lee2020learning} is an inherently discrete approach and cannot be trained to output continuous actions, nor can it be trained on data with actions outside of its discrete action space.

\hlc{
\mbox{\tbl{table:20k_results}} shows that the performance of all Lee \textit{et al.} variants is poor compared to FFN, particularly on 1-step goals (see \appfig{fig:sim_lee} and \appfig{fig:sim_lee_ms_results} for qualitative results).
FFN outperforms Lee \textit{et al.} when trained on either the discrete action dataset or the continuous one.
Training FFN on continuous actions results in better performance for 1-step goals, but the discrete action dataset also performs fairly well. 
These results indicate that the improved performance of FFN vs. Lee \textit{et al.} cannot be solely explained by training on continuous vs. discrete action data, though other factors like outputting continuous actions instead of discrete ones may still play significant role in FFN's improved performance.}

\begin{table*}[ht!]
    \centering
    \caption{\hl{Mean Particle Distance Error for Lee~\mbox{\etal} on 20k Training Examples}}
    \label{table:20k_results}
    \begin{tabular*}{0.95\textwidth}{l @{\extracolsep{\fill}} d{2.2} @{\extracolsep{\fill}} d{2.2} @{\extracolsep{\fill}} d{2.2} }
      \toprule
        Baseline 
            & \multicolumn{1}{c}{1-Step (40)} 
            & \multicolumn{1}{c}{Multi Step (6)} 
            & \multicolumn{1}{c}{All (46)} \\
        \midrule
Lee \textit{et al.}, 1-Arm, D, No CB, LC & 18.94,16.43 & 24.18,17.75 & 19.62,16.49 \\
Lee \textit{et al.}, 1-Arm, D, No CB, HC & 16.18,08.38 & 26.20,16.31 & 17.49,10.10 \\
Lee \textit{et al.}, 1-Arm, D, CB, LC & 20.99,18.88 & 34.61,31.35 & 22.77,20.97 \\
Lee \textit{et al.}, 1-Arm, D, CB, HC & 19.70,09.37 & 38.91,24.05 & 22.20,13.53 \\
Lee \textit{et al.}, 1-Arm, RGB, No CB, LC & 49.29,18.10 & 52.03,33.62 & 49.65,20.26 \\
Lee \textit{et al.}, 1-Arm, RGB, No CB, HC & 47.12,21.04 & 64.48,29.85 & 49.38,22.75 \\
Lee \textit{et al.}, 1-Arm, RGB, CB, LC & 33.89,19.01 & 58.90,43.34 & 37.15,24.38 \\
Lee \textit{et al.}, 1-Arm, RGB, CB, HC & 39.01,25.36 & 55.46,38.38 & 41.15,27.43 \\
Lee \textit{et al.}, 2-Arm, D, No CB, LC & 36.62,14.51 & 47.72,21.95 & 38.07,15.82 \\
Lee \textit{et al.}, 2-Arm, D, No CB, HC & 40.75,13.22 & 52.88,19.03 & 42.33,14.45 \\
Lee \textit{et al.}, 2-Arm, D, CB, LC & 47.18,18.60 & 57.29,28.65 & 48.50,20.07 \\
Lee \textit{et al.}, 2-Arm, D, CB, HC & 35.98,24.60 & 64.75,51.76 & 39.73,30.30 \\
        \hlc{FFN, 2-Arm, D, CB, HC, Discrete Actions} &
         9.57,06.07 & \mathbf{10.15},\mathbf{07.20} & 10.17,07.34\\
        \midrule
        FFN, 2-Arm, D, CB, HC, Cont. (Ours) & \mathbf{4.46},\mathbf{02.62} & 25.04,22.88 & \mathbf{7.14},\mathbf{11.06}\\
        \bottomrule
    \end{tabular*}
    \vspace{2pt}
    \caption*{\footnotesize D: Depth \quad CB: Corner Bias \quad LC: Low Camera \quad HC: High Camera \quad Cont: Continuous Actions}
    \vspace{-1em}
\end{table*}

\hlc{\mbox{\textbf{Lee \textit{et al.} with and without Subgoals.}}} \hl{FFN uses subgoals at inference time in order to fully specify the task; many cloth folding goals have final goal configurations in which large portions of the cloth are self-occluded. Subgoals are required to ensure the task is completed correctly and that the cloth is correctly folded. Lee~\mbox{\etal\cite{lee2020learning}} demonstrated cloth folding without subgoals at inference time by relying on a learned Q-value heatmap to select actions toward a final end goal. We compare the performance of the best Lee~\mbox{\etal} variant with and without subgoals at test-time. The results of this experiment are in \mbox{\tbl{table:subgoals}}.
While the performance on 1-step goals are similar because those tasks do not have subgoals, performance on multi-step goals is worse without subgoals.}

\begin{table*}[ht!]
    \centering
    \caption{\hl{Mean Particle Distance Error for Lee~\mbox{\etal} With and Without Subgoals}}
    \label{table:subgoals}
    \begin{tabular*}{0.9\textwidth}{l @{\extracolsep{\fill}} d{2.2} @{\extracolsep{\fill}} d{2.2} @{\extracolsep{\fill}} d{2.2} }
      \toprule
        Method 
            & \multicolumn{1}{c}{1-Step (40)} 
            & \multicolumn{1}{c}{Multi Step (6)} 
            & \multicolumn{1}{c}{All (46)} \\
        \midrule
Lee \textit{et al.} & 16.92,9.28 & 37.74,38.99 & 19.71,20.27 \\
Lee \textit{et al.}, With Subgoals & \mathbf{16.18},\mathbf{8.38} & \mathbf{26.20},\mathbf{16.31} & \mathbf{17.49},\mathbf{10.10} \\
        \bottomrule
    \end{tabular*}
\end{table*}

\section{Additional Details and Results for Ablations}
\label{sec:app_ablations}

\subsection{Ablation Implementation Details}

\textbf{NoFlowIn} The architecture for this ablation is identical to our main method, except that it takes depth images instead of flow images as input. We use a conditioned architecture with two PickNets; PickNet1 receives the observation and goal depth images as input both of size $200 \times 200$. The place point is computed by querying the flow image similar to our main method.

\textbf{NoFlowPlace} We predict the place points similarly to the pick points by using an additional place network. The place network architecture is identical to PickNet. The input is a flow image and the output is the place point predictions.

\textbf{NoFlow} This ablation is a combination of NoFlowIn and NoFlowPlace, where PickNet and PlaceNet both take observation and goal depth images as input.

\textbf{NoCornerBias} This ablation is the same as our main method except for the training dataset. We use a dataset that does not bias the data to pick corners (See \sect{sec:ffn-implementation-details}). 
Instead, the pick actions are always uniformly sampled over the visible cloth mask. 
We still constrain the folding actions for both arms to be in the same direction and distance from their respective pick points and point towards the center of the frame.

\textbf{NoSplitPickNet} The architecture of PickNet is modified so that we only have one PickNet for both arms instead of the conditioned architecture used in our main method. The PickNet takes as input the flow image and outputs two heatmaps corresponding to the two pick points.

\textbf{NoMinLoss} The loss in Eq. 1 is replaced with the following:

\begin{equation}
   \begin{aligned}
  \mathcal{L}_\text{NoMin} = \text{BCE}(H_1, H_1^*) + \text{BCE}(H_2, H_2^*)
  \label{eq:lnomin}
  \end{aligned}
\end{equation}

\subsection{Additional Ablation Results} 

We provide ablation results in \tbl{table:more_sim_results} grouped by single-step, multi-step, and all goals.

\begin{table*}[ht!]
    \centering
    \caption{\hl{Mean Particle Distance Error for Ablations}}
    \label{table:more_sim_results}
    \normalsize
    \begin{tabular*}{0.75\textwidth}{l @{\extracolsep{\fill}} d{2.2} @{\extracolsep{\fill}} d{2.2} @{\extracolsep{\fill}} d{4.4} }
      \toprule
        Ablation 
            & \multicolumn{1}{c}{One Step (n=40)} 
            & \multicolumn{1}{c}{Multi Step (n=6)} 
            & \multicolumn{1}{c}{All (n=46)} \\
        \midrule
NoFlowIn & 5.14,3.62 & 24.63,21.30 & 9.37,12.20 \\
NoFlowPlace & 7.61,5.44 & 30.25,17.62 & 10.56,11.15 \\
NoFlow & 8.97,7.45 & 28.79,19.33 & 18.02,20.34 \\
NoCornerBias & 9.79,5.57 & \mathbf{19.61},\mathbf{17.52} & 11.07,8.83 \\
NoSplitPickNet & 4.87,2.61 & 23.41,18.87 & 7.29,9.56 \\
NoMinLoss & 5.10,4.04 & 20.81,17.57 & 7.15,9.08 \\
        \midrule
        FFN (Ours) & \mathbf{4.46},\mathbf{02.62} & 25.04,22.88 & \mathbf{7.14},\mathbf{11.06} \\
        \bottomrule
    \end{tabular*}
\end{table*}

\section{Additional Results on Unseen Cloth Shapes}
\begin{figure}[htb!]
    \centering
    \begin{subfigure}[t]{0.22\textwidth}
        \includegraphics[width=\textwidth, trim={0 1em 0 0},clip]{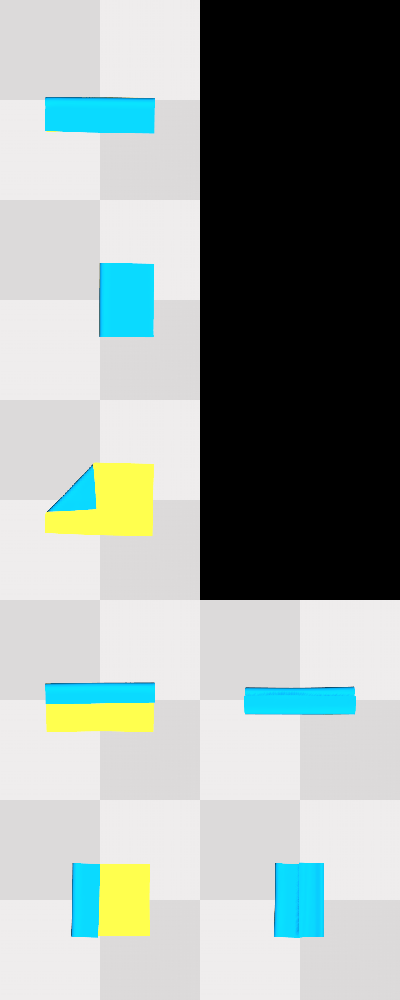}
        \caption{Rect. cloth goals}
        \label{fig:rect_goals_sim}
    \end{subfigure}
    \begin{subfigure}[t]{0.22\textwidth}
        \includegraphics[width=\textwidth, trim={0 1em 0 0},clip]{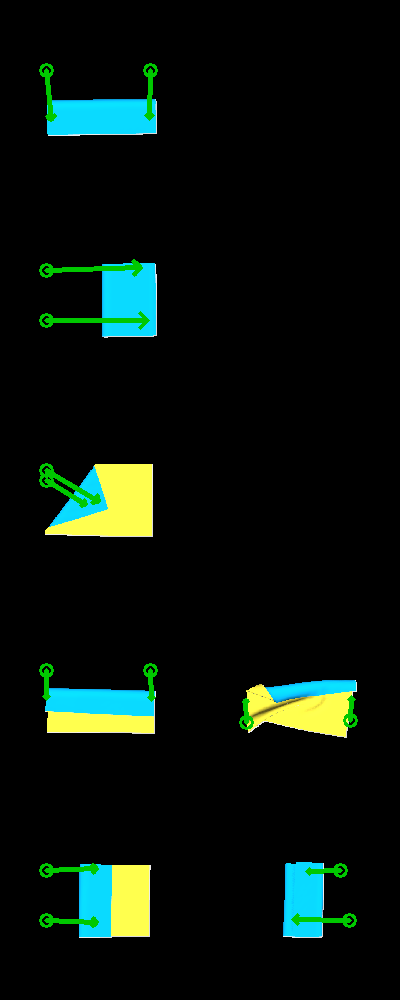}
        \caption{FFN achieved}
        \label{fig:rect_sim_ffn}
    \end{subfigure}
    \begin{subfigure}[t]{0.22\textwidth}
        \includegraphics[width=\textwidth, trim={0 1em 0 0},clip]{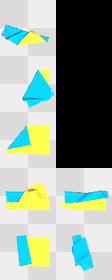}
        \caption{Fabric-VSF achieved}
        \label{fig:rect_vsf}
    \end{subfigure}
    \begin{subfigure}[t]{0.22\textwidth}
        \includegraphics[width=\textwidth, trim={0 1em 0 0},clip]{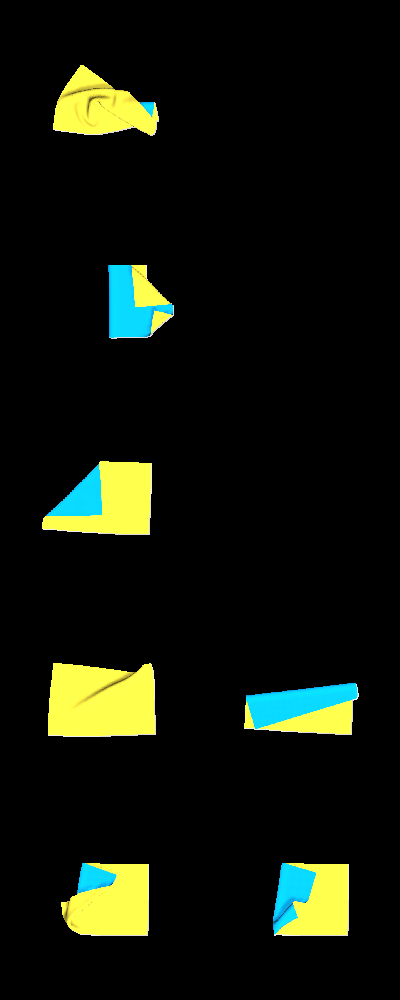}
        \caption{Lee~\textit{et al.} achieved}
        \label{fig:rect_lee}
    \end{subfigure}
    \caption{Qualitative performance of FFN, Fabric-VSF, and Lee~\textit{et al.} on rectangular cloth.}
    \label{fig:rect_cloth}
    % \vspace{-0.5em}
\end{figure}

We also evaluate Fabric-VSF \hl{and Lee~\mbox{\etal}} on generalization to unseen cloth shapes.
FFN generalizes well to new shapes, as shown in the main text (see Fig.~5 and Sec.~4.2.1).
\tbl{table:sim_generalization} provides quantitative results on the rectangle cloth and T-shirt for the best Fabric-VSF method and best Lee~\mbox{\etal\ }method compared to FFN. 
FFN outperforms both methods by a large margin.
Fabric-VSF generalizes poorly, likely because it relies on planning with a learned visual dynamics model.
Lee\mbox{~\etal\ }also does not generalize well compared to FFN.
 \fig{fig:rect_cloth} provides a qualitative comparison. 

\begin{table*}[ht!]
    \centering
    \caption{\hl{Mean Particle Distance for Folding Unseen Cloth Shapes in Simulation}}
    \label{table:sim_generalization}
    \normalsize
    \begin{tabular}{l d{2.2} d{2.2}}
      \toprule
        Method 
            & \multicolumn{1}{c}{Rectangle (n=6)} 
            & \multicolumn{1}{c}{T-Shirt (n=3)} \\
        \midrule
        Lee~\etal, 1-Arm, No Corner Bias, High Cam, 20k Actions & 31.63,18.04 & 86.65,34.67 \\
        Fabric-VSF, 1-Arm, Corner Bias, Large Action & 25.68,11.21 & 45.25,13.83 \\
        FFN (Ours) & \mathbf{10.70},\mathbf{08.54} & \mathbf{20.91},\mathbf{11.28} \\
        \bottomrule
    \end{tabular}
    % \vspace{5pt}
\end{table*}

\section{End-to-End Variants of FFN}

\hl{We investigate the effect of training our FFN architecture end-to-end. First, we train the FFN architecture with pick losses as well as the flow loss; all losses are allowed to backpropagate through the entire combined network, including through the FlowNet layers. The results on the square towel are presented in \mbox{\tbl{table:e2e_results}} (``JointFFN''). This variant performs significantly worse than FFN (9.28 vs. 7.14 on all goals). 
}

\begin{table*}[htb!]
    \centering
    \caption{Mean Particle Distance Error (mm) for End-to-End Variants of FFN}
    \label{table:e2e_results}
    \normalsize
    \begin{tabular}{l d{2.2} d{2.2} d{2.2}}
      \toprule
        Method
            & \multicolumn{1}{c}{1-Step (n=40)} 
            & \multicolumn{1}{c}{Multi-Step (n=6)} 
            & \multicolumn{1}{c}{All (n=46)} \\
        \midrule
        JointFFN  & 07.60,05.62 & \mathbf{17.53},\mathbf{15.56} & 09.28,09.39  \\
        JointPredictPlace  & 12.90,11.67 & 35.25,19.22 & 22.88,23.24 \\
        JointFFN, No Flow Loss  & 32.41,22.61 & 68.17,50.35 & 37.07,30.34 \\
        JointPredictPlace, No Flow Loss & 16.31,22.73 & 50.27,31.44 & 24.39,29.77 \\
        \midrule
        FFN (Ours) & \mathbf{4.46},\mathbf{02.62} & 25.04,22.88 & \mathbf{7.14},\mathbf{11.06} \\
        \bottomrule
    \end{tabular}
\end{table*}

\hl{We also trained another variant which consists of a FlowNet, a PickNet, and a PlaceNet, trained end-to-end (``JointPredictPlace'' in \mbox{\tbl{table:e2e_results}}). This is similar to our ablation ``PredictPlace'' in \mbox{\tbl{table:ablations}}, which uses the same architecture but is not trained end-to-end. JointPredictPlace performs significantly worse than FFN (22.88 vs. 7.14 on all goals) and also underperforms compared to PredictPlace (10.56 on all goals). Overall, this result, as well as the one in the paragraph above, indicate that end-to-end training leads to significantly worse performance for this task. Our intuition for this is that the flow network should be trained only with the flow loss, and that backpropagating the gradients from the pick loss into the flow network adds noise and reduces its performance.}

\hl{Lastly, we evaluated variants of the above two architectures with the flow loss removed, to see if we could train these architectures end-to-end with just a single loss at the end, instead of using an intermediate flow loss. The results, shown in \mbox{\tbl{table:e2e_results}}, are worse for both variants, showing the importance of the intermediate flow loss.  
}

\section{FFN Performance with Crumpled Starting Configurations}

\begin{figure}[ht!]
    \centering
    \begin{subfigure}[t]{0.20\textwidth}
        \includegraphics[width=\textwidth, trim={0 1em 0 0},clip]{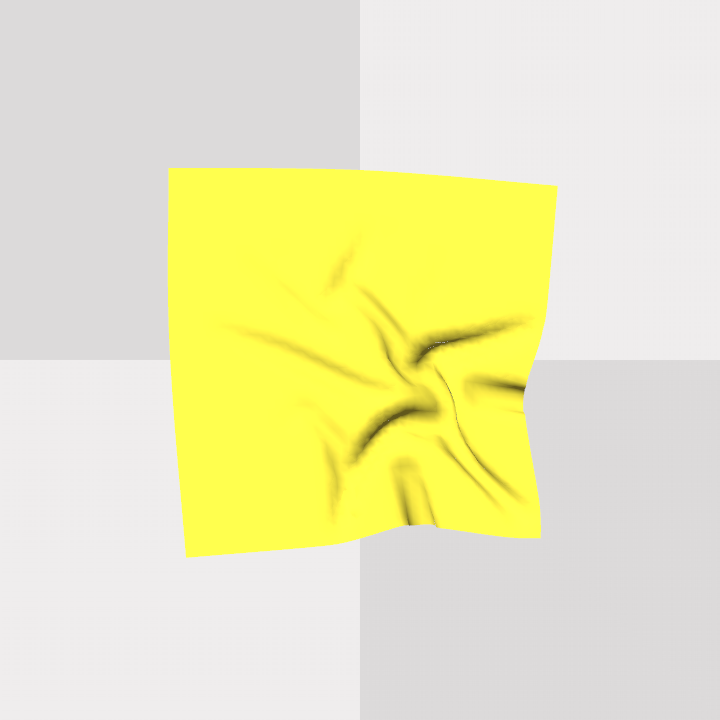}
        \caption{Crumpled 0}
    \end{subfigure}
    \begin{subfigure}[t]{0.20\textwidth}
        \includegraphics[width=\textwidth, trim={0 1em 0 0},clip]{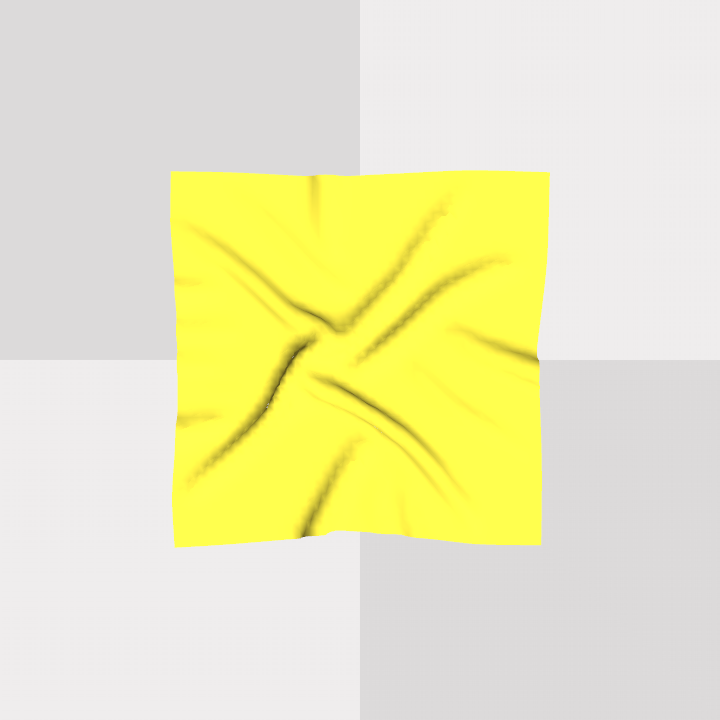}
        \caption{Crumpled 1}
    \end{subfigure}
    \begin{subfigure}[t]{0.20\textwidth}
        \includegraphics[width=\textwidth, trim={0 1em 0 0},clip]{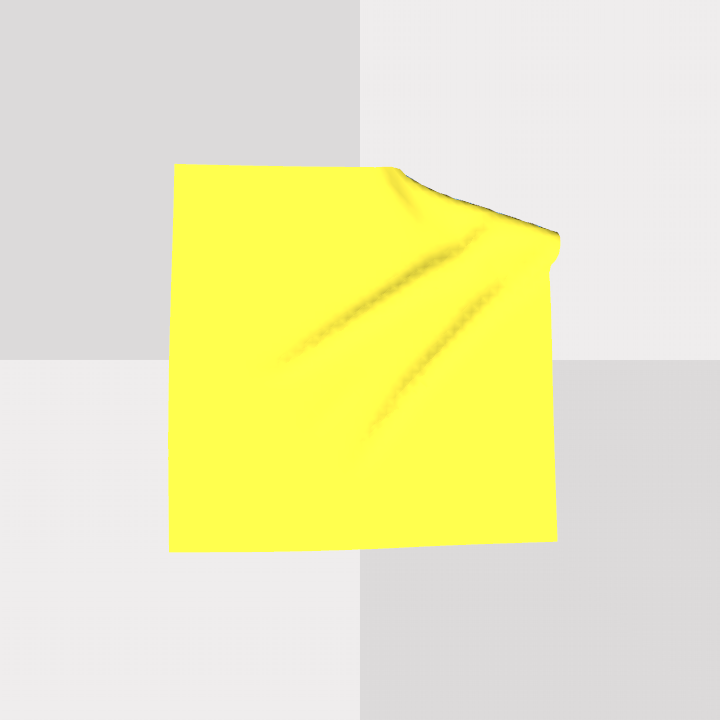}
        \caption{Crumpled 2}
    \end{subfigure}
    \caption{\hl{Crumpled initial cloth configurations}} 
    \label{fig:crumple_cloth}
\end{figure}

\hl{Our experiments focused on folding tasks, and we assume that a previous method was used to flatten the cloth before our method is executed. To evaluate the robustness of our method to imperfect smoothing, we evaluate the performance of FFN in simulation on slightly crumpled initial cloth configurations. We generated crumpled configurations by taking the flat cloth and executing a random pick and place action with a maximum translation of 10 pixels. The three configurations used in our experiments are shown in \mbox{\fig{fig:crumple_cloth}}.}

\begin{figure}[ht!]
    \centering
    \begin{subfigure}[t]{0.65\textwidth}
        \includegraphics[width=\textwidth, trim={0 1em 0 0},clip]{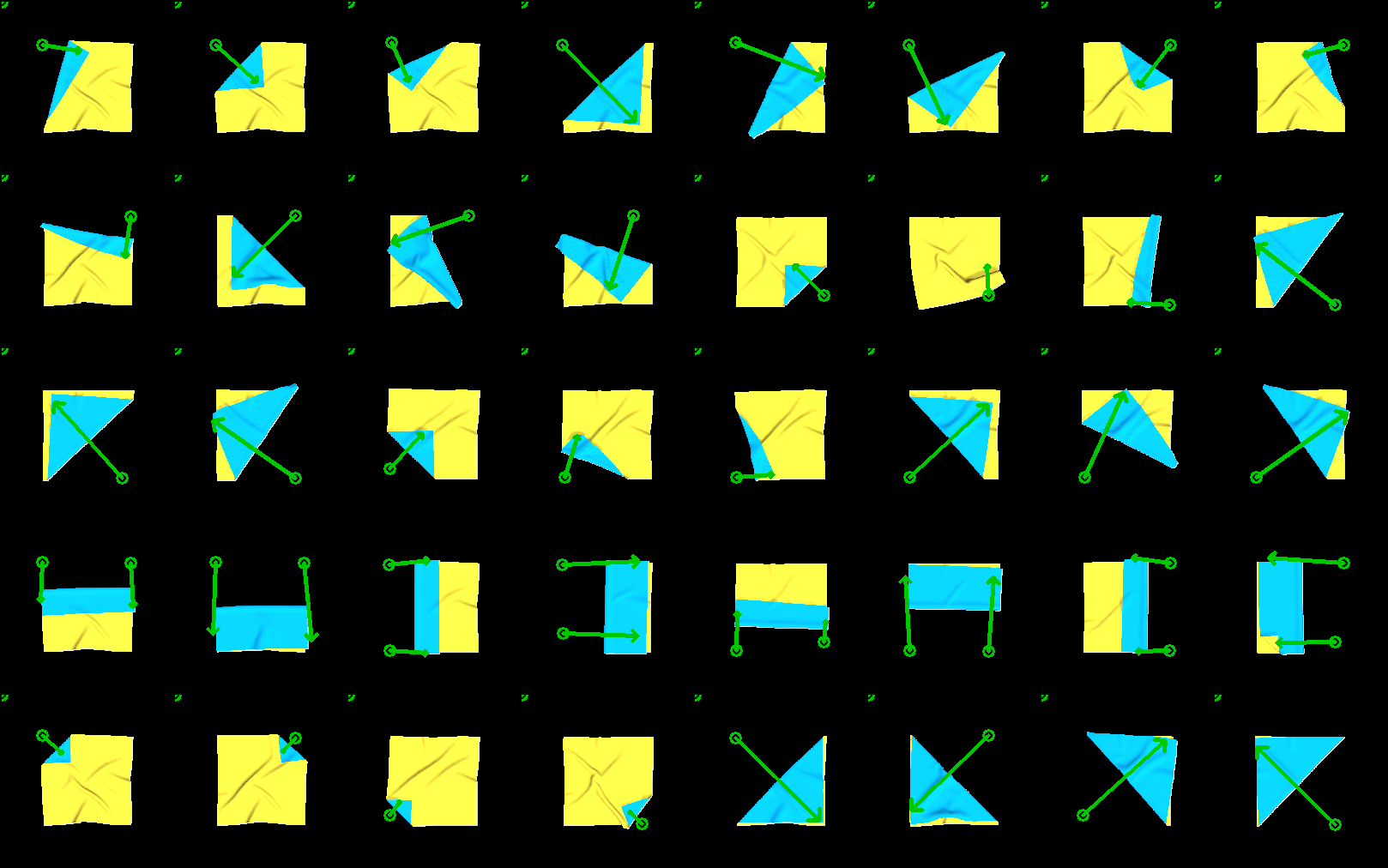}
        \caption{Crumpled one-step FFN performance}
    \end{subfigure}
    \begin{subfigure}[t]{0.32\textwidth}
        \includegraphics[width=\textwidth, trim={0 1em 0 0},clip]{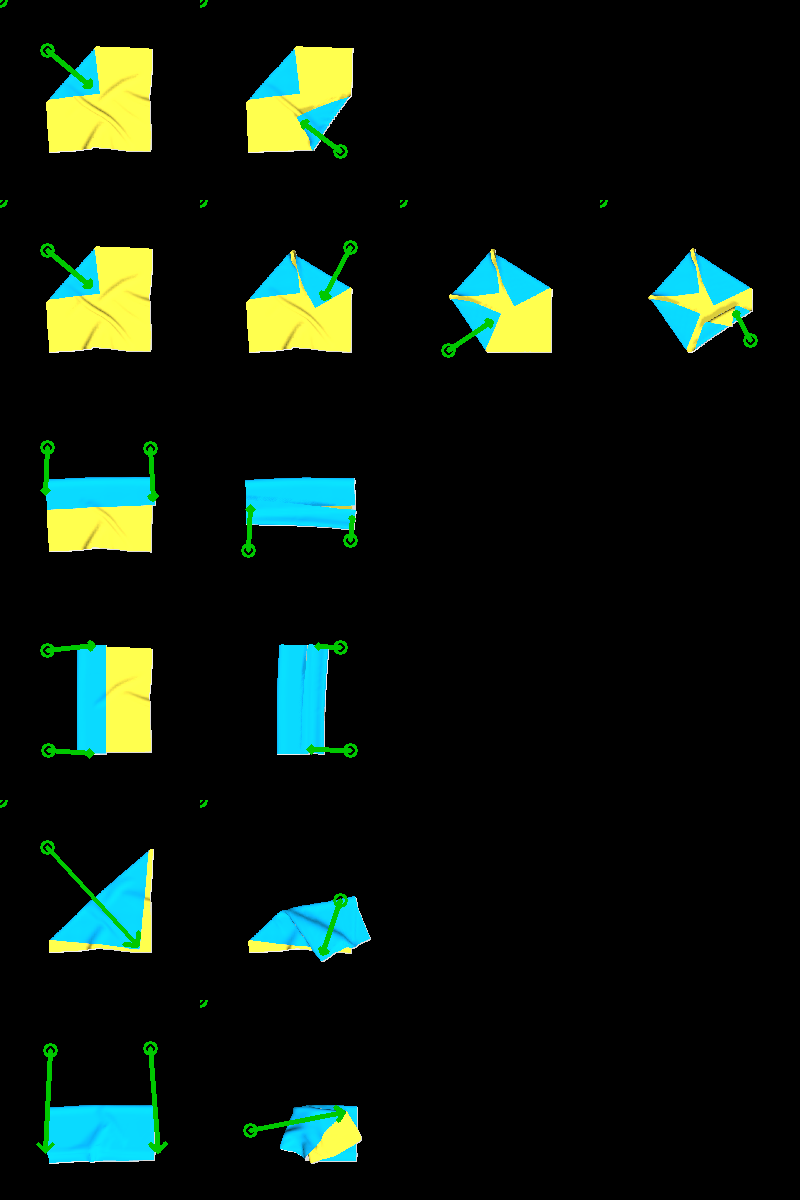}
        \caption{Crumpled Multi-step FFN performance}
    \end{subfigure}
    \caption{\hl{Configurations achieved by FFN when starting from the ``Crumpled 1'' configuration for each attempt (compare with} \fig{fig:sim_goals_results})} 
    \label{fig:crumple_achieved}
\end{figure}

\hl{For each crumpled configuration, we evaluated FFN on the full set of 46 evaluation goals, where the starting configuration of the cloth was set to the given crumpled configuration. The results of these evaluations are in \mbox{\tbl{table:crumple_results}}. The particle distance error is slightly higher with the crumpled starting configurations, but the qualitative results in \mbox{\fig{fig:crumple_achieved}} show that FFN still produces actions that are very close to the intended goals.
}

\begin{table*}[ht!]
    \centering
    \caption{\hl{Mean Particle Distance Error (mm) for FFN with Different Start Configurations}}
    \label{table:crumple_results}
    % \normalsize
    \begin{tabular}{l d{4.4} d{4.4} d{4.4}}
      \toprule
        Starting Config
            & \multicolumn{1}{c}{1-Step (n=40)} 
            & \multicolumn{1}{c}{Multi-Step (n=6)} 
            & \multicolumn{1}{c}{All (n=46)} \\
        \midrule
        FFN, Crumpled 0  & 12.40,4.82 & 24.82,24.81 & 14.01,10.86  \\
        FFN, Crumpled 1  & 10.68,2.89 & 23.54,22.56 & 12.36,9.61 \\
        FFN, Crumpled 2  & 10.68,4.29 & \mathbf{21.05},\mathbf{14.70} & 12.03,7.51 \\
        FFN, Flat & \mathbf{4.46},\mathbf{2.62} & 25.04,22.88 & \mathbf{7.14},\mathbf{11.06}   \\
        \bottomrule
    \end{tabular}
    % \vspace{2pt}
    % \caption*{\footnotesize CB: Corner Bias \quad Sm. Action: Small Action \quad Lg. Action: Large Action}
\end{table*}

\section{FFN Performance with Iterative Refinement}

\begin{table*}[ht!]
    \centering
    \caption{\hlc{Mean Particle Distance Error (mm) for FFN with Iterative Refinement}}
    \label{table:iter_refine}
    % \normalsize
    \begin{tabular}{l d{4.4} d{4.4} d{4.4}}
      \toprule
        Starting Config
            & \multicolumn{1}{c}{1-Step (n=40)}
            & \multicolumn{1}{c}{Multi-Step (n=6)} 
            & \multicolumn{1}{c}{All (n=46)} \\
        \midrule
        FFN, No Refinement & \mathbf{4.46},\mathbf{2.62} & 25.04,22.88 & 7.14,11.06   \\
        FFN, Iterative Refinement  & 4.54,2.58 & \mathbf{20.47},\mathbf{19.49} & \mathbf{6.62},\mathbf{9.17} \\
        \bottomrule
    \end{tabular}
\end{table*}

\hlc{In our normal evaluations, each goal or subgoal is attempted only once by each method. 
With a single attempted action for each subgoal, FFN is able to achieve a diverse set of goals, as demonstrated in this work. 
However, we find that FFN can achieve even better performance when attempting goals multiple times, using the flow to compare the current observation with the goal and taking actions that move the observation closer to the goal if it has not yet been reached.
We evaluate the benefit of using this ``iterative refinement'' procedure in simulation. 
% \textbf{\textcolor{red}{
FFN moves to the next subgoal when a minimum threshold for the average flow is achieved, so the flow acts as a goal recognizer.
The policy is allowed a maximum of 3 iterative actions per subgoal to limit potential divergence.
% }}
The results in \mbox{\tbl{table:iter_refine}} show that iterative refinement can improve performance, particularly on multi-step goals, where reaching the current subgoal accurately is important for achieving subsequent goals. }

\section{FlowNet Performance}

\begin{figure}[hbt!]
    \centering
    \begin{subfigure}[t]{0.99\textwidth}
        \includegraphics[width=\textwidth]{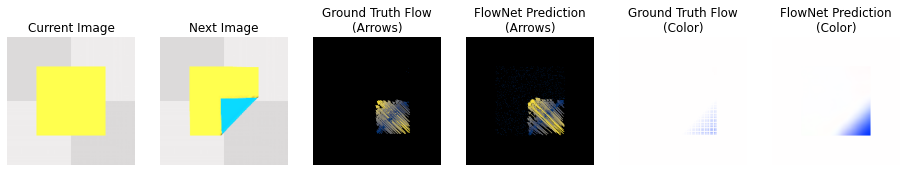}
    \end{subfigure}
    \begin{subfigure}[t]{0.99\textwidth}
        \includegraphics[width=\textwidth]{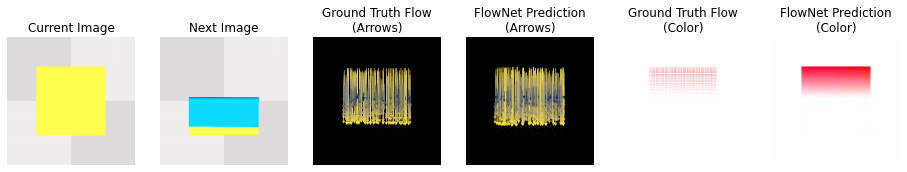}
    \end{subfigure}
    \caption{FlowNet Qualitative Performance. Two types of visualizations are provided: representing the flow vector as arrows, and representing the flow vector using RGB channels. FlowNet outputs a dense flow image but is trained on sparse ground truth flow. FlowNet takes only depth images as input; RGB images are shown as a visual aid only.}
    \label{fig:flownet_perf}
\end{figure}

FlowNet achieves an average endpoint error (EPE) of 1.0268 on the set of simulated test goals.
The test goals are not seen during training. 
\fig{fig:flownet_perf} provides qualitative examples of FlowNet performance on simulated test goals. 

\newpage
\putbib[supplement/appendix]
\end{bibunit}
\end{document}